\documentclass[12pt, a4paper]{article}
\usepackage[T1]{fontenc}
\usepackage{tgbonum}
\usepackage{times}
\usepackage{amsmath, amssymb, amsfonts}
\usepackage{graphicx}
\usepackage{epstopdf}
\usepackage{caption, subcaption}
\usepackage[margin=01 in]{geometry}
\usepackage[onehalfspacing]{setspace}
\usepackage{xr-hyper}
\usepackage{hyperref}
\usepackage[affil-it]{authblk}
\numberwithin{equation}{section}
\usepackage{amsthm}

\usepackage{authblk}
\usepackage{url}
\usepackage{cite}
\usepackage{xurl}
\usepackage[table,xcdraw]{xcolor}
\usepackage{lineno} 
\usepackage{float}
\usepackage{tcolorbox}
\usepackage{orcidlink}

\usepackage{array}
\newcolumntype{C}[1]{>{\centering\let\newline\\\arraybackslash\hspace{0pt}}m{#1}}
\usepackage{siunitx} 

\providecommand{\keywords}[1]
{
  \small	
  \textbf{\textit{Keywords---}} #1
}

\title{An early warning indicator trained on stochastic disease-spreading models with different noises} 

\author[1]{Amit K. Chakraborty\orcidlink{0000-0002-7960-3984}} 
\author[1]{Shan Gao\orcidlink{0009-0002-1705-8033}}
\author[2]{Reza Miry}
\author[2]{Pouria Ramazi\orcidlink{0000-0003-4906-0090}}
\author[3,4]{Russell Greiner}
\author[5]{Mark A. Lewis\orcidlink{0000-0002-7155-7426}}
\author[1,*]{Hao Wang\orcidlink{0000-0002-4132-6109}}
\affil[1]{Department of Mathematical and Statistical Sciences, University of Alberta, AB, Canada}
\affil[2]{Department of Mathematics and Statistics, Brock University, ON, Canada}
\affil[3]{Department of Computing Science, University of Alberta, AB, Canada}
\affil[4]{Alberta Machine Intelligence Institute, AB, Canada}
\affil[5]{Department of Mathematics and Statistics and Department of Biology, University of Victoria, BC, Canada}
\affil[*]{Corresponding author: hao8@ualberta.ca}
\date{}
\begin{document}


\maketitle

\begin{abstract}
The timely detection of disease outbreaks through reliable early warning signals (EWSs) is indispensable for effective public health mitigation strategies. Nevertheless, the intricate dynamics of real-world disease spread, often influenced by diverse sources of noise and limited data in the early stages of outbreaks, pose a significant challenge in developing reliable EWSs, as the performance of existing indicators varies with extrinsic and intrinsic noises. Here, we address the challenge of modeling disease when the measurements are corrupted by additive white noise, multiplicative environmental noise, and demographic noise into a standard epidemic mathematical model. To navigate the complexities introduced by these noise sources, we employ a deep learning algorithm that provides EWS in infectious disease outbreak by training on noise-induced disease-spreading models. The indicator's effectiveness is demonstrated through its application to real-world COVID-19 cases in Edmonton and simulated time series derived from diverse disease spread models affected by noise. Notably, the indicator captures an impending transition in a time series of disease outbreaks and outperforms existing indicators. This study contributes to advancing early warning capabilities by addressing the intricate dynamics inherent in real-world disease spread, presenting a promising avenue for enhancing public health preparedness and response efforts. 
\end{abstract} 


\normalsize	

\section{Introduction}
Infectious diseases have imposed significant burdens on societies, and their impact has intensified in recent decades \cite{baker2022infectious, sweileh2022global}. While public health measures have made considerable progress in eradicating certain diseases, the continuous emergence of new diseases and the re-emergence of some older ones have resulted in substantial loss of life and economic hardship \cite{southall2021early}. Over the past two decades, more than ten severe infectious diseases have (re-)emerged, resulting in approximately 7.3 million global deaths \cite{baker2022infectious, bloom2019infectious, lepan2020visualizing, whoCOVID, world2018managing}. With infectious diseases persistently on the rise at a global level, the necessity for an effective early warning signal (EWS) becomes increasingly imperative \cite{kamalrathne2023need}. 

An EWS is a model-independent tool that anticipates an impending critical transition of a disease outbreak before it escalates to an epidemic level, without relying on empirically fitted mathematical models. \cite{southall2021early}. This allows for the early identification of outbreaks, which, in turn, can potentially help in reducing the spread of the disease and minimizing its impact through targeted interventions \cite{yang2017early, guo2020early, languon2019filovirus}. Although an effective EWS has the potential to save millions of lives from future diseases, designing an efficient early warning system that can adapt to the complex and uncertain patterns often exhibited by real-world disease dynamics remains a significant challenge \cite{proverbio2022performance}. 

EWSs are diverse and can be generated using a wide range of detection methods, including wastewater-based epidemiology, search engine trends, and human bio-monitoring \cite{sims2020future}. While these methods are increasingly popular, their implementation may require additional resources and can differ greatly from one country to another \cite{sims2020future}. For instance, wastewater-based epidemiology necessitates the selection of appropriate biomarkers, search engine trends require internet access with precise keyword configurations, and human bio-monitoring demands the careful selection of a control group \cite{sims2020future}.

Critical slowing down (CSD) serves as the fundamental basis for EWSs to anticipate critical transitions \cite{southall2021early}. As a system approaches a critical threshold or undergoes a significant change, such as a phase transition, its ability to recover from perturbations decreases \cite{brett2017anticipating}. This slowing down results in an increase in recovery time and a decrease in the recovery rate from small perturbations \cite{scheffer2012anticipating, bury2021deep}. Mathematically, CSD occurs due to the diminishing real part of the dominant eigenvalue as a system approaches the bifurcation point, reaching zero precisely at the bifurcation point \cite{brett2017anticipating}, where bifurcation refers to qualitative changes in the behavior of a dynamical system's steady state as its parameters are varied \cite{strogatz2018nonlinear}. Detection of critical slowing down in a system can be achieved by measuring temporal trends in statistics, such as increased autocorrelation (AC), variance, and the magnitude of fluctuations \cite{scheffer2012anticipating, southall2021early}. 

The reliability of CSD as an indicator of an EWS is not guaranteed when a system involves stochasticity \cite{scheffer2012anticipating, boettiger2013early}. In highly stochastic scenarios, the dynamics of a system can accelerate rather than slow down near a bifurcation point \cite{titus2020critical}, and a transition often occurs far away from the local bifurcation point \cite{liu2015identifying, scheffer2012anticipating, titus2020critical}. The leading statistical measures, such as variance and AC, are differentially affected due to the external and internal effects of noise, and exhibit different trends for different types of noises \cite{titus2020critical}. Traditional eigenvalue-based analyses in dynamical systems are also ineffective in predicting state changes in stochastic systems, as unpredictable fluctuations can significantly distort signals generated by linear terms \cite{liu2015identifying}. As stochasticity leads to unforeseen system behavior, the deterministic framework of the underlying mechanisms becomes obscured, potentially resulting in false alarms or delayed responses.

The impact of noise is a critical consideration when generating EWS for disease outbreaks, as the data observed from real-world systems is convoluted with different types of noises \cite{liu2015identifying}. Various forms of randomness can be introduced in a system by both intrinsic and extrinsic noises \cite{o2018stochasticity}. Two common forms of noise used when modeling disease are environmental and demographic noise. Environmental noise, also known as extrinsic noise, arises due to the changes in environmental factors outside of population \cite{allen2010introduction, boettiger2018noise}. It plays a crucial role in zoonotic, vector-borne, and waterborne diseases, particularly in Ebola, avian influenza, malaria, and cholera transmission \cite{allen2017primer, carmona2020winter}. On the other hand, demographic noise, referred to as intrinsic noise, is associated with temporal variation within the population process, including births, deaths, immigration, emigration, and state transitions \cite{allen2010introduction,constable2016demographic}. In addition to these two types of noise, additive white noise represents the interplay between deterministic structures and random fluctuations with consistent intensity, possessing the potential to trigger an epidemic in circumstances where there was no initial spread of infection in the absence of this specific noise \cite{tuckwell2000enhancement}. Since EWSs arise from the interaction of the dynamics and the noise, the type of noise will affect the nature and intensity of an EWS. 

To study the data-driven approach of EWS in infectious diseases, many use generic early warning indicators (EWIs), such as variance and AC \cite{drake2019statistics, dessavre2019problem, o2016leading, o2018stochasticity, southall2020prospects, o2019disentangling, proverbio2022performance, brett2020detecting, o2013theory, brett2018anticipating}, and machine learning models \cite{deb2022machine, brett2020dynamical, dylewsky2023universal, bury2021deep}. While generic EWIs offer a computationally efficient method, their pattern can fluctuate, depending on system behavior and noise levels. Moreover, the performance of generic EWIs is sensitive to the data sampling, quantitative and objective measures, length of the time series, and the size of window and bandwidth \cite{deb2022machine}. This introduces uncertainty regarding their universal applicability or potential influence by specific dynamic features \cite{proverbio2022performance, southall2021early, boettiger2013early}. In addition to these challenges, generic EWIs can not accurately predict the type of bifurcation that is going to happen \cite{bury2021deep}. Machine learning models, in contrast to generic EWIs, can not only identify the impending change in the system but also categorize the type of bifurcation expected according to the training labels \cite{bury2021deep}. These models possess the capacity to recognize and learn patterns, allowing them to capture the statistical features inherent in the data \cite{deb2022machine}. Nevertheless, machine learners, particularly deep learners (DL), demand substantial amounts of training data to produce effective models and are computationally intensive. 

Training machine learning models with synthetic data is a fundamental strategy, especially valuable when collecting and labeling extensive real datasets is challenging, time-consuming, resource-intensive, and restricted due to privacy concerns \cite{nikolenko2021synthetic}.  Synthetic data generation involves the creation of artificial datasets that closely mimic real-world data, which can produce better models as it involves greater data diversity \cite{nikolenko2021synthetic}. One of the key advantages of synthetic data lies in the control it provides over data characteristics \cite{rankin2020reliability, nikolenko2021synthetic}. This control is invaluable for researchers and practitioners, enabling the exploration of a wide array of scenarios, data distributions, and complexities \cite{rankin2020reliability}. Moreover, synthetic data facilitates model training on edge cases and rare events, which may occur in the real world but only rarely \cite{jacobsen2023machine}. This becomes particularly crucial in applications where the consequences of failure are substantial or the process is highly intricate, such as in medical diagnostics or disease spread \cite{rankin2020reliability}. Synthetic scenarios capturing these exceptional cases empower models to make more informed decisions when confronted with unusual situations, ultimately enhancing the model's robustness and reliability. 

A recent study conducted by Bury et al. \cite{bury2021deep} demonstrated that a machine learning model can outperform generic EWIs, providing more accurate EWS by analyzing the pre-transition time series of a system. They employed a DL algorithm trained on a synthetic dataset generated from two-dimensional random dynamical systems comprising a general polynomial of up to third order. The parameter values of the random dynamical system were adjusted to achieve fold, Hopf, and transcritical bifurcations, and the stochastic simulations with additive white noise were then conducted to generate training data. Their study suggested that by training the DL algorithm on a diverse range of datasets, it could capture fundamental features shared by various systems, thus enabling it to provide EWS for systems not explicitly included in the training data. The model was tested on time series from ecology, thermoacoustics, climatology, and epidemiology, which were not part of the training datasets, and successfully anticipated critical transitions in both simulated models and empirical data. Moreover, it accurately identified the specific types of bifurcations, that were occurring within these systems. Nevertheless, when applied to a disease-spreading model, both the DL model and generic EWIs did not perform better than random chance. This motivated us to investigate the performance of EWIs in the context of a disease model. 

Our objective is to enhance EWSs for infectious diseases. To achieve this, we evaluate the effectiveness of a DL algorithm, alongside variance and lag-1 AC, in disease-spreading models. To comprehensively analyze the interaction between disease dynamics and noise, we consider a standard epidemic model with additive white noise, multiplicative environmental noise, and demographic noise. Our hypothesis posits that by training the DL algorithm on noise-induced models, the algorithm can outperform other EWIs and demonstrate higher accuracy in providing EWSs for disease outbreaks. We use the time series of the number of infected cases to generate EWSs. We specifically focus on the transcritical form of bifurcation in disease models, identifying the bifurcation point by observing the basic reproduction number ($R_0$)--a dimensionless number that denotes the average number of secondary infections arising from a single infected case in a fully susceptible population. A disease model can exhibit other types of bifurcation \cite{gumel2012causes}, which are not in our consideration.

\section{Mathematical models and the impacts of noises} 
We consider the susceptible–infectious–recovered (SIR) model \cite{kermack1927contribution}, which serves as a foundation for most disease-spreading models. The SIR model divides the population into susceptible $(S)$, infectious $(I)$, and recovered $(R)$ compartments and uses a system of ordinary differential equations to describe how individuals move between them over time. The model equations are given by
\begin{equation}
\begin{cases}
\frac{dS(t)}{dt} = \Lambda -\beta(t) S(t)I(t)-\mu S(t), \\
\frac{dI(t)}{dt} = \beta(t) S(t)I(t)-\alpha I(t) -\mu I(t), \\
\frac{dR(t)}{dt} = \alpha I(t) -\mu R(t),
\end{cases}
\label{SIR_model_pre}
\end{equation}
where $\Lambda$ is the recruitment rate of susceptible populations, $\beta(t)$ is the time-dependent disease transmission rate, $\mu$ is the natural death rate, and $\alpha$ is the recovery rate from the disease. When $\beta$ is treated as a constant parameter, the disease-free equilibrium point of the model \eqref{SIR_model_pre} is $(\Lambda, 0, 0)$, the endemic equilibrium point is $(\frac{\mu+\alpha}{\beta}, \frac{\Lambda}{\mu+\alpha}-\frac{\mu}{\beta}, \frac{\alpha\Lambda}{\mu(\mu+\alpha)}-\frac{\alpha}{\beta})$, and the basic reproduction number, $R_0$, is $\frac{\beta \Lambda}{\mu(\alpha+\mu)}$. A disease goes extinct if $R_0<1$ and persists if $R_0>1$, and a transcritical bifurcation occurs when $R_0=1$, where both equilibria meet and exchange their stability \cite{gumel2012causes, proverbio2022performance, southall2021early}. The bifurcation value, corresponding to $R_0 = 1$, arises from the model with a constant $\beta$ which is regarded as a bifurcation parameter.  
The first two equations of \eqref{SIR_model_pre} are independent of $R$, and the reduced system is given by
\begin{equation}
\begin{cases}
\frac{dS(t)}{dt} &= \Lambda -\beta(t) S(t)I(t)-\mu S(t), \\
\frac{dI(t)}{dt} &= \beta(t) S(t) I(t)-\alpha I(t) -\mu I(t).
\end{cases}
\label{SIR_model_reduced}
\end{equation}


To model the spread of an infectious disease by an exposed individual, the SIR-type model is commonly expanded by including an exposed compartment, leading to the susceptible–exposed-infectious–recovered (SEIR) model \cite{brauer2008mathematical}. The model equations are given by 
\begin{equation}
\begin{cases}
\frac{d S(t)}{d t} = \Lambda - \beta(t) S(t) I(t) -d S(t), \\
\frac{d E(t)}{d t} = \beta(t) S(t) I(t)-(d+\kappa) E(t), \\
\frac{d I(t)}{d t} = \kappa E(t)-(d+\gamma) I(t) , \\
\frac{d R(t)}{d t} = \gamma I(t)-d R(t), 
\end{cases}
\label{SEIR_model}
\end{equation}
where $S$, $E$, $I$, and $R$ denote the susceptible, exposed, infectious, and recovered individuals respectively. Moreover, $\Lambda$ is the recruitment rate of the susceptible population, $\beta(t)$ is the disease transmission rate, $d$ is the natural death rate, $1/ \gamma$ is the mean infectious period, and $1/ \kappa$ is the mean exposed period. The basic reproduction number of the SEIR model is $\frac{\kappa \beta \Lambda}{d (d+\kappa) (d+\gamma)}$, and the bifurcation point is $\beta_c = \frac{d (d+\kappa) (d+\gamma)} {\kappa \Lambda}$.

\subsection {Noise-induced models}

Deterministic models provide useful insights into disease dynamics under idealized conditions. However, the dynamics of infectious diseases are inherently complex and subject to random fluctuations, thus necessitating the careful consideration of noise in disease models. Noise disrupts the deterministic structure and influences the dynamics of the population compartments. These noises can arise from various sources, including diversity in population behavior, environmental factors, and even the stochastic nature of disease transmission. A noise-induced model is commonly defined through a set of stochastic differential equations.

\subsubsection{SIR and SEIR models with additive white noise}

The most common noise model is the additive noise model, wherein a perturbed observed compartment is construed as the sum of unaltered deterministic dynamics and an additive noise component \cite{bovik2009basic}. This noise model serves as a means to simulate the inherent stochasticity of natural processes and reproduce their consequential impact on a system \cite{hutt2022additive}. To model additive white noise, we add a white noise term into each equation of the deterministic model, encompassing the continuous intensity of fluctuation associated with the dynamics of individual compartments. A stochastic differential equation with additive white noise is given by 
\begin{equation}
    dX(t) = f(X(t),t)dt + \sigma dW(t),  
\end{equation}
where $X(t)$ is a state variable, $\sigma$ is noise intensity and $W(t)$ is a Wiener process. Following this, the SIR model \eqref{SIR_model_reduced} with additive white noise has the form  
\begin{eqnarray}\label{WhiteNoise_Model}
\begin{cases}
\displaystyle{dS(t)} = \displaystyle{ \Lambda dt -\beta(t) S(t)I(t) dt-\mu S(t) dt+\sigma_1 dW_1(t) }, \\
\displaystyle{dI(t)} = \displaystyle{\beta(t) S(t)I(t) dt-\alpha I(t) dt -\mu I(t) dt+\sigma_2 dW_2(t)},\\
\end{cases}
\end{eqnarray}
where $\sigma_i,$ $i = 1, 2,$  are the intensities of white noise and $W_i(t),$ $i = 1, 2,$ are independent Wiener processes. Similarly, the SEIR model \eqref{SEIR_model} with additive white noise is given by

\begin{equation}
\begin{cases}
dS(t) = \Lambda dt - \beta(t) S(t) I(t) dt -d S(t) dt + \sigma_1 dW_1(t), \\
dE(t) = \beta(t) S(t) I(t) dt-(d+\kappa) E(t) dt + \sigma_2 dW_2(t), \\
dI(t) = \kappa E(t) dt-(d+\gamma) I(t) dt + \sigma_3 dW_3(t), \\
dR(t) = \gamma I(t) dt -d R(t) dt + \sigma_4 dW_4(t), 
\end{cases}
\label{SEIR_whiteN}
\end{equation}
where $\sigma_i,$ $i = 1, 2, 3, 4,$  are the intensities of white noise and $W_i(t),$ $i = 1, 2, 3, 4,$ are independent Wiener processes.

\subsubsection{SIR model with multiplicative environmental noise}

Substantial environmental noise can trigger abrupt transitions in disease dynamics even before reaching a critical threshold; nevertheless, in certain instances, environmental fluctuations can alter the stability landscape and promote resilience \cite{majumder2021finite}. Environmental variability introduces multiplicative noise into the modeling process that depends on the system's current state \cite{liu2017stationary}. To model the multiplicative environmental noise, we consider that the fluctuations exhibit a direct correlation with the size of the population compartment. Consequently, we incorporate a multiplication by the population compartment in the noise-induced terms within the dynamics of the respective compartments. A stochastic differential equation with multiplicative environmental noise is given by  
\begin{equation}
    dX(t) = f(X(t),t)dt + \sigma X(t) dW(t).  
\end{equation}
Therefore, following \eqref{SIR_model_reduced}, the SIR model with multiplicative environmental noise is given by 
\begin{eqnarray}\label{EnvNoise_model}
\begin{cases}
\displaystyle{dS(t)} = \displaystyle{ \Lambda dt -\beta(t) S(t)I(t) dt-\mu S(t) dt+ \sigma_1 S(t) dW_1(t)},\\
\displaystyle{dI(t)} = \displaystyle{\beta(t) S(t)I(t) dt-\alpha I(t) dt-\mu I(t) dt+\sigma_2 I(t) dW_2(t)}.\\
\end{cases}
\end{eqnarray} 
 
\subsubsection{SIR model with demographic noise}
Demographic noise can either mitigate abrupt transitions or promote alternative stable states and abrupt transitions \cite{majumder2021finite}. Moreover, demographic noise has some other epidemiological outcomes such as early decrease of infection, epidemic outbreaks with the collapse of population, and either the extinction of infection or an endemic situation \cite{lambert2018demographic}. To characterize demographic stochasticity, we use a system of stochastic differential equations that is based on a diffusion process where both the time and the state variables are continuous. Following the derivation procedure in \cite{allen2010introduction, allen2017primer}, we incorporate demographic noise into the model \eqref{SIR_model_reduced}. Examining the interactions within the SIR model \eqref{SIR_model_reduced}, we identify the list of five possible changes along with their respective probabilities, as presented in Table \ref{demographic-transition-probability}.   
\begin{table}[hbt!]
\footnotesize
\caption{Transition probabilities associated with changes in model \ref{SIR_model_reduced}}
\centering 
\begin{tabular}{|>{\columncolor{gray!25}}C{1 cm}|C{4 cm}|C{4 cm}|C{5 cm}|} 
\hline
\rowcolor{gray!25}
$i$ & Description & Change $(\Delta X)_i$ \ & Probability, $p_i$ \\ 
\hline %
1 & Recruitment of S & (1, 0)$^{tr}$ & 
$\Lambda \Delta t$ \\
\hline
2 & Transmission from S to I & (-1, 1)$^{tr}$ & 
$\beta S I\Delta t$ \\
\hline
3 & Death of S  & (-1, 0)$^{tr}$ & 
$\mu S \Delta t$  \\
\hline
4 & Death of I & (0, -1)$^{tr}$ & 
$\mu I \Delta t$ \\
\hline
5 & Recovery of I & (0, -1)$^{tr}$   & 
$\alpha I \Delta t$ \\
\hline 
6 & No change & (0, 0)$^{tr}$ & 
$1-(\Lambda+\beta S I +\mu S + \mu I + \alpha I) \Delta t$ \\
\hline
\end{tabular}
\label{demographic-transition-probability} 
\end{table} 
To incorporate demographic stochasticity, we need to calculate the expectation $E(\Delta X)$ and the covariance $\sum(\Delta X)$ for the time interval $\Delta t$. The expectation is given by 
\[E(\Delta X) = \sum_{i=1}^{6} p_i(\Delta X)_i = \begin{pmatrix}
\Lambda-\beta SI-\mu S \\
\beta SI-\alpha I-\mu I
\end{pmatrix} 
\Delta t = 
\mu \Delta t. \]
The covariance is given by
\begin{align*}
    E((\Delta X) (\Delta X)^{tr}) = 
    \sum_{i=1}^{6} p_i(\Delta X)_i (\Delta X)_i^{tr} & = \begin{pmatrix}
    \Lambda+\beta SI+\mu S & -\beta SI \\
    -\beta SI&\beta SI+\alpha I+\mu I
    \end{pmatrix}
    \Delta t \\
    & =
    \begin{pmatrix}
    v_{11} & v_{12} \\
    v_{21} & v_{22}
    \end{pmatrix}
    \Delta t = V \Delta t.
\end{align*}
Letting $\Delta t \rightarrow 0$, then the stochastic SIR model with demographic noise has the form
\[dX = \mu dt + G dW(t),\]
where $X=(S, I)^{tr}$, and $G$ is the square root of $V$ defined by
\[G =\sqrt{V}=
\frac{1}{e}
\begin{pmatrix}
v_{11} + d & v_{12} \\
v_{21} & v_{22} + d
\end{pmatrix} \]
where $ v_{12}=  v_{21}$, $d = \sqrt{v_{11} v_{22} - v_{12}^2}$, and $e = \sqrt{v_{11} + v_{22} + 2d}$. Choosing, $v_{11} = a$, $ v_{12} = v_{21} = b$, $v_{22} = c$, and substituting $d$ and $e$ in the $G$ matrix, we get the following stochastic SIR model with demographic noise 
\begin{eqnarray}\label{DemNoise_Model}
\begin{cases}
\displaystyle{dS(t)} = \displaystyle{\Lambda dt -\beta(t) S(t)I(t) dt-\mu S(t) dt+ \frac{a(t)+d(t)}{e(t)} dW_1(t)+\frac {b(t)}{e(t)} dW_2(t)}, \\
\displaystyle{dI(t)} = \displaystyle{\beta(t) S(t)I(t) dt-\alpha I(t) dt - \mu I(t) dt + \frac{b(t)}{e(t)} dW_1(t) + \frac{c(t)+d(t)}{e(t)} dW_2(t)}, \\ 
\end{cases}
\end{eqnarray}
where
$a(t) = \displaystyle{\Lambda +\beta(t) S(t)I(t) +\mu S(t)}$,  
$b(t) = -\beta(t) S(t) I(t)$, 
$c(t) = \beta(t) S(t)I(t) + \alpha I(t) + \mu I(t)$, 
$d(t) = \sqrt{a(t) c(t) - b^2(t)}$, 
$e(t) =\sqrt{a(t)+c(t)+2d(t)}$, 
and $W_i(t),$ $i = 1, 2,$ is the Wiener process.

\section{Methods}
\subsection{Simulated training data} \label{training data}
To train the DL algorithm, we generated the training data from the noise-induced SIR models, i.e., from equations \eqref{WhiteNoise_Model}, \eqref{EnvNoise_model}, and \eqref{DemNoise_Model}. The training data comprises two types of simulations: \textit{transcritical} and \textit{null}. In transcritical simulations, bifurcation occurs within a specified time length, while in null simulations there is no bifurcation within the same time frame. The maximum time series length is set to 1500, and we assume that the transition time for the noise-induced models aligns with the deterministic model \eqref{SIR_model_reduced}. Specifically, the transition in the noise-induced models occurs when the value of $R_0$ becomes equal to $1$. Therefore, when we use data, we obtain the critical value of the disease transmission rate, $\beta_c$, at the bifurcation point by setting $R_0 = 1$, resulting in $ \beta_c = \frac{\mu (\alpha + \mu)}{\Lambda}$.

\begin{figure}[tb!]
\centering
 \begin{subfigure}[t]{0.31\textwidth}
    \includegraphics[width=\linewidth]{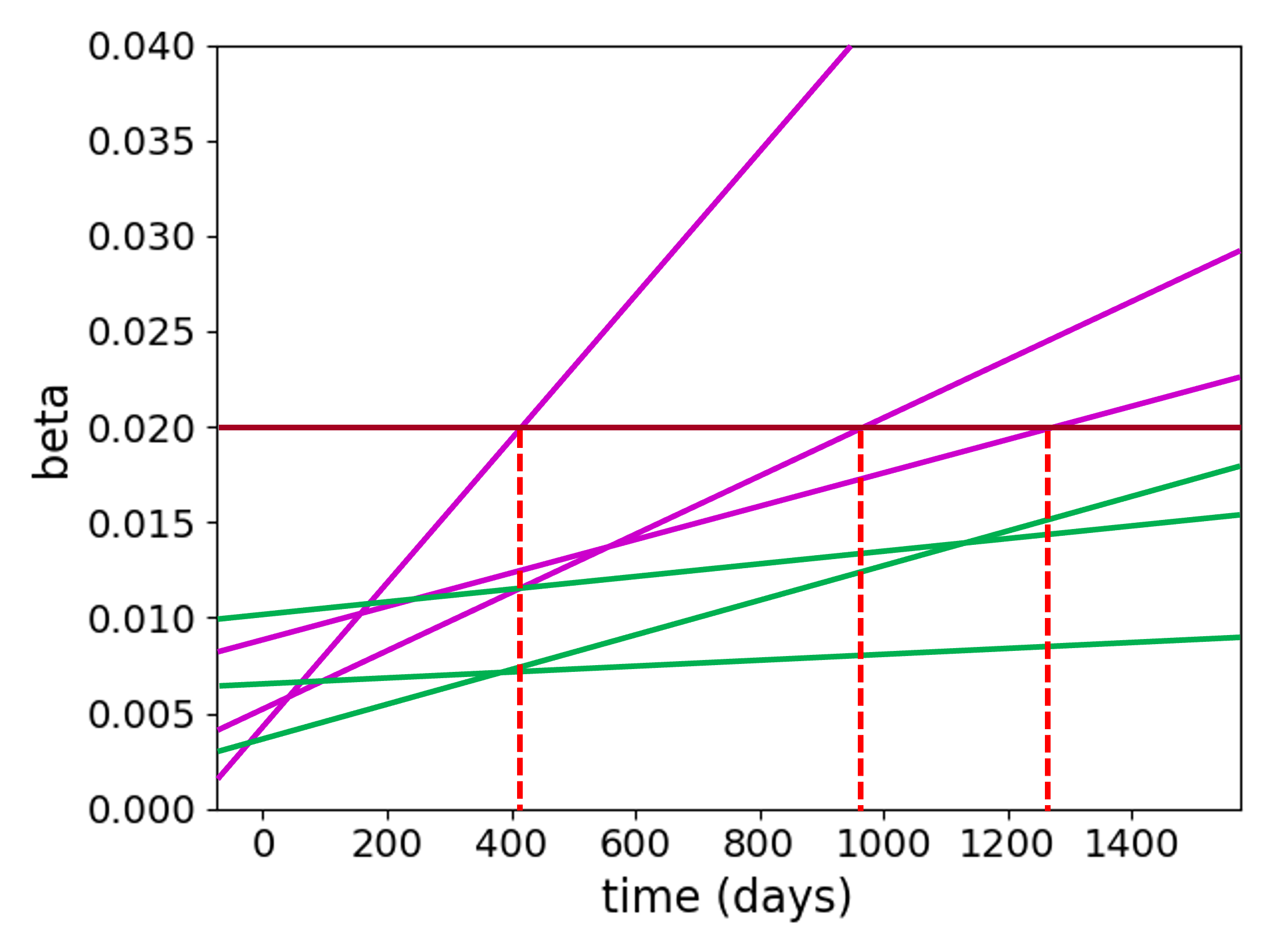}
    \put(-155,100){\textbf{(a)}}
 \end{subfigure}
 \hfill
  \begin{subfigure}[t]{0.32\textwidth}
    \includegraphics[width=\linewidth]{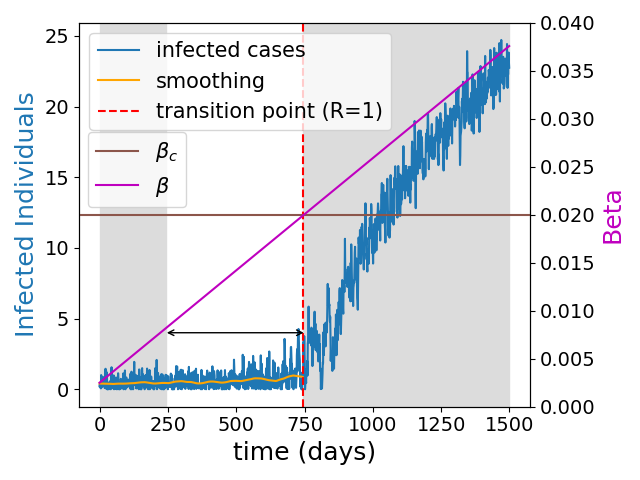}
    \put(-153,100){\textbf{(b)}}
 \end{subfigure}
  \hfill
  \begin{subfigure}[t]{0.32\textwidth}
    \includegraphics[width=\textwidth]{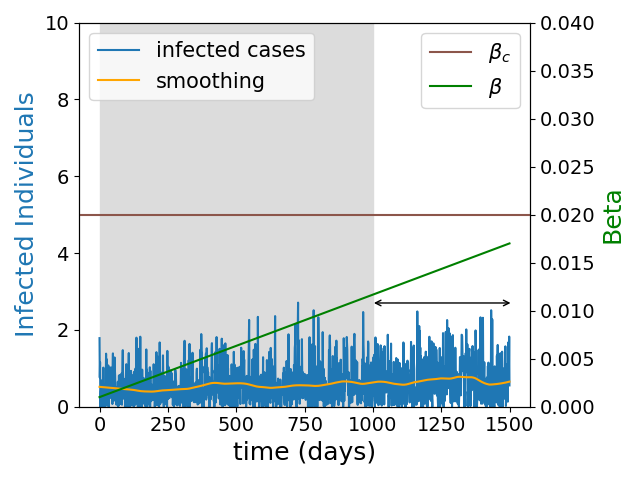}
    \put(-153,100){\textbf{(c)}}
 \end{subfigure}
\caption{(a) Linear increase samples of $\beta(t)$ for null (green) and transcritical (purple) simulations. A transition occurs when $\beta$ crosses the critical value $\beta_c$ (brown horizontal line), with the transition time marked by the intersection of the two lines (red dotted vertical lines). In transcritical simulations, $\beta$ crosses $\beta_c$ randomly between time 0 and 1500, while in null simulations, $\beta$ does not cross $\beta_c$ before time 1500. (b) If a transition occurs between time 0 and 1500, the number of infected of the preceding 500 (100) points leading up to the bifurcation point are utilized as training data for transcritical simulations. (c) In the absence of a transition during the period from time 0 to 1500, the number of infected of the most recent 500 (100) time points are selected as training data for null simulations.}
\label{inc_of_beta}
\end{figure}

We specify the parameters $\Lambda = 100$, $\alpha = 1$, and $\mu = 1$, and randomly choose other parameters to incorporate variability into the training data, ensuring that the DL algorithm is trained on a wide range of scenarios. The disease transmission rate, $\beta$, is assumed to be time-dependent, following a linear increasing pattern: $\beta(t) = \beta_0 + \beta_1 t$, where $\beta_0$ and $\beta_1$ are chosen from triangular distributions. For $\beta_0$, the lower and upper limits are set to $0$ and $\frac{\beta_c}{2}$ respectively. The choice of $\beta_1$ determines whether a simulation is transcritical or null. For null simulations, $\beta_1$ ranges from $0$ (lower limit) to $\frac{\beta_c - \beta_0}{1500}$ (upper limit), whereas for transcritical simulations, $\beta_1$ ranges from $\frac{\beta_c - \beta_0}{1500}$ (lower limit) to $\frac{2 \times \beta_c - \beta_0}{1500}$ (upper limit). The modes of these distributions are set at half of their respective upper limits. These parameters result in a random transition between time 0 and 1500 for transcritical simulations and no transition for null simulations (Figure \ref{inc_of_beta}). Random transition times are chosen for transcritical simulations to account for the uncertain nature of transition times in real-world scenarios. These parameter choices collectively contribute to the model's robustness by exposing it to a wide range of conditions, fostering adaptability, and enhancing its potential to generalize effectively across different scenarios and uncertainties.

The Wiener process is sampled from a normal distribution, $dW_i \sim \mathcal{N}(0, \sqrt{dt}), i= 1,2,$ and the noise intensity, $\sigma_i,$ $ i= 1,2,$ is sampled from a triangular distribution with a lower limit of $0$, a mode of $0.5$, and an upper limit of $1$, and is randomized for each simulation. The random choice of noise intensity introduces different levels of stochasticity into the simulations, making them more adaptable to real-world scenarios where accurately measuring all parameters and noises is often impractical. These variations provide a broader spectrum of possible phenomena.

We set the initial values of susceptible and infected to 500 and 7 respectively, with a burn-in period of 100 time units. The Euler–Maruyama method is employed for numerical simulations with $dt = 0.01$. For the transcritical simulations, simulation lengths extend up to the transition points, whereas for the null simulations, the length extends up to time 1500. We generated 30,000 simulations from each of the equations \eqref{WhiteNoise_Model}, \eqref{EnvNoise_model} and \eqref{DemNoise_Model}, and an additional 10,000 simulations from equation \eqref{WhiteNoise_Model} by doubling the noise intensity. Thus, a total of 0.1 million simulations were used to train the DL algorithm. 

The training data was generated in multiple batches using the compute Canada server. Time series residuals were computed using the \texttt{ewstools} Python package \cite{bury2023ewstools}, utilizing Lowess (Locally Weighted Scatterplot Smoothing) smoothing \cite{cleveland1979robust} with a span of 0.2. Additionally, lag-1 AC and variance of the simulations were computed with a rolling window of 0.25 using the same package.

\subsection{Deep learning algorithm and training}

We assume that the occurrence or absence of an upcoming transition can be reliably determined by examining recent time points within the time series and we define two subsets of observations of each time series: one is 100 time points, and another is 500 time points. Therefore, for transcritical simulations, a subset of observations consists of the preceding 500 (100) time points before a bifurcation point, and for null simulations, it consists of the last 500 (100) time points (Figure \ref{inc_of_beta}). If a transition occurs before the 500th (100th) time point, the values beyond the transition point are padded with zeros. This ensures that the transition point is always included in the dataset for transcritical simulations. 

We followed the same DL algorithm, training process, and hyperparameters as outlined in \cite{bury2021deep}. The neural network architecture combined Convolutional Neural Networks (CNNs) and Long Short-Term Memory (LSTM) layers \cite{mutegeki2020cnn, vidal2020gold, bury2021deep}. Since we have a two-class classification problem, we modified the output (dense) layer of the neural network to 2 units with a softmax activation function. 

The DL algorithm was trained on two subsets of observations, i.e., lengths of 500 and 100 points. The residuals of the time series were used to train the algorithm, normalized by dividing each individual value of a time point by the mean absolute value of the entire residuals. The algorithm underwent 1500 epochs of training, and the learning rate was set at 0.0005. Hyperparameters underwent tuning through a series of grid searches, and the training, test, and validation split was 95\%, 1\%, and 4\%, respectively. The F1 score, precision, and recall for an ensemble of two models for the 500-length trained were all 97.6\%, and for the 100-length trained were all 91.0 \%. The average prediction at each data point was taken from two trained models of each length (either 500 or 100) and an ensemble was formed.

\subsection{Testing}
To assess the performance of the DL model, we conducted tests using four noise-induced disease-spreading mathematical models: three corresponding to the models used during training, and the fourth model featuring higher dimensionality compared to our trained models. Additionally, real-world empirical data on COVID-19 cases in Edmonton was included in the test.

\subsubsection{Mathematical models}
To test the SIR model with additive white noise \eqref{WhiteNoise_Model}, multiplicative environmental noise \eqref{EnvNoise_model}, and demographic noise \eqref{DemNoise_Model}, we simulated time series using the same parameters and configurations employed during the training phase. Due to the inherent randomness of certain parameters, these time series may exhibit variations compared to the training simulations. Our objective is to determine whether the DL model can accurately classify the transition regardless of these random variations.

To evaluate the predictive performance of the DL model for a higher dimensional model than the trained model, we test the SEIR model with additive white noise \eqref{SEIR_whiteN}. We use parameter values $\Lambda = 100$, $d = 0.75$, $\kappa = 2$, $\gamma = 1$, $S_0 = 500$, $E_0 = 1$, $I_0 = 2$, and $R_0 = 2$, to simulate the model. The disease transmission rate and other parameters remain the same as those chosen for generating the training data discussed in section \ref{training data}.

We generated 20 time series from each test model, comprising 10 transcritical simulations and 10 null simulations. In each transcritical simulation, the time series extended up to the bifurcation point between time 0 and 1500, while each null simulation extended up to time 1500. These simulated time series served as input for trained models to identify the presence of a bifurcation, thereby facilitating the evaluation of the DL model's performance.

\subsubsection{Empirical data}
We used the COVID-19 dataset of Edmonton sourced from \cite{COVID-19Edmonton}, covering daily cases from March 2020 to July 2023. Employing the \texttt{EpiEstim} package \cite{cori2020package, nash2023estimating} in the R programming language, we computed the effective reproduction number ($R_{e}$) from the reported incidence cases in biweekly sliding windows. This package requires an estimated serial interval distribution of the disease and we considered the mean of the serial interval of COVID-19 to be 6.3, with a standard deviation of 4.2 \cite{nash2023estimating}. 

We extracted consecutive daily case data from the entire time series based on periods where the mean $R_{e}$ was less than 1, up to the point where $R_{e}$ reached 1 (Supplementary Figures~\ref{empirical_covid_Re}). This process resulted in obtaining sub-series of varying lengths. To ensure consistency, we selected sub-series with a minimum length of eight weeks. Following this criterion, we derived a total of 7 sliced time series immediately preceding the bifurcation point, which we labeled as transcritical time series. For null time series, we adjusted the length of each transcritical time series by shortening it by four weeks. This adjustment makes the null time series data up to four weeks before the bifurcation point. Our objective is to assess the capability of the trained DL model to detect the impending transition based on these pre-transition time series.

\subsection{Prediction and ROC curve}

We employed the DL algorithm trained on two distinct types of data to generate EWSs for both simulated noise-induced time series datasets and empirical data of COVID-19. First, we utilized the 500-classifier DL algorithm from Bury et al. \cite{bury2021deep}, which was trained on data generated from random dynamical systems with polynomial terms, which we will call the PODATR model. Secondly, we employed the DL algorithm trained on data from noise-induced SIR models of time lengths 100 and 500 which we will call the SIDATR-100 and SIDATR-500 models respectively. Of course, we want an accurate classification of both the occurrence and non-occurrence of a transition. Consequently, the trained models assign probabilities for each type of bifurcation in the training set. Thus, based on inputs of the number of infected from a subset of observations, the SIDATR-100 and SIDATR-500 models assign probabilities for both transcritical and null bifurcations, while the PODATR model assigns probabilities for each of fold, Hopf, transcritical, and null bifurcations. A type of bifurcation with the highest probability assigned at the end of the input observation period is considered as a predictive signal for the occurrence of that bifurcation type.

Our prediction process follows the methodology outlined in Bury et al. \cite{bury2021deep}. We used an expanding window of ten time points for predictions, resulting in a maximum of 50 (10) predictions from a time series spanning up to 500 (100) points. We extracted a set of the last 5 predictions from each of the time series, allowing for a comprehensive assessment of model performance when it reaches the bifurcation point. Thus, for each type of simulation from the mathematical models, we generated a maximum of 50 (10) predictions to ensure a robust and well-balanced analysis. 

We compared the performance of the DL models to the lag-1 AC and variance, using the area under the receiver operating characteristic (ROC) curve. The ROC curve illustrates the relationship between sensitivity (true positive rate) and 1 – specificity (false positive rate) across varying discrimination thresholds, offering a comprehensive view of the model's overall performance. An ideal warning system would detect all thresholds without any false alarms, resulting in an area under the ROC curve equal to one. Conversely, an ineffective system would struggle to differentiate between false and true signals.

To assess generic EWIs, variance and lag-1 AC, we use the Kendall $\tau$ value using \texttt{scipy.stats} package in Python. This statistic serves as an indicator of either an increasing or decreasing trend, where higher and positive values indicate a more strongly increasing trend \cite{bury2021deep}. Consequently, we employed these indicators to make predictions for a specific outcome when the Kendall $\tau$ statistic surpassed a designated discrimination threshold \cite{bury2021deep}.

\section{Results}

\subsection{Performance on mathematical models}
\begin{figure}[!tbh]
    \centering
    \includegraphics[width=0.95\linewidth]{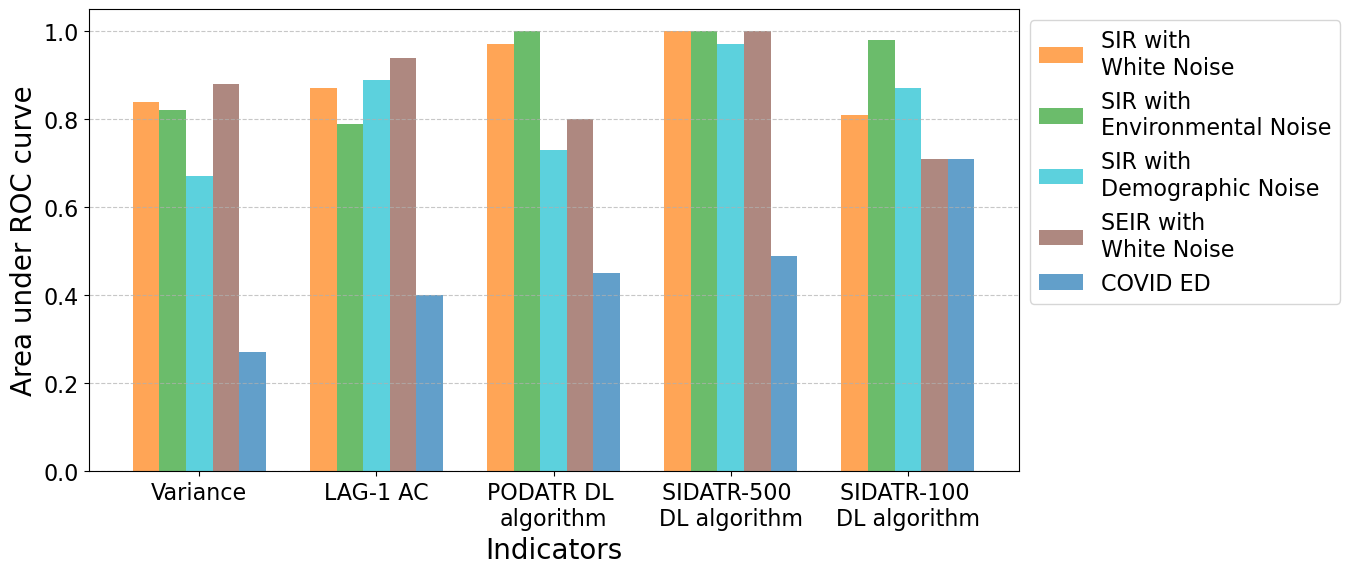}
    \caption{Area under the ROC curve of the generic EWIs--variance, and lag-1 AC--in addition to the PODATR, SIDATR-500 and SIDATR-100 DL model. Performance was assessed on the last five predictions of transcritical and null simulations of the SIR model with white noise (yellow), SIR model with multiplicative environmental noise (green), SIR model with demographic noise (cyan), SEIR model with white noise (brown), and COVID-19 dataset of Edmonton (blue).}
    \label{AUC-bar}
\end{figure}

The area under the ROC curve (AUC) for lag-1 AC exceeded 0.85 in the simulations of the SIR model with white noise, the demographic noise, and the SEIR model, outperforming variance (Figure \ref{AUC-bar}). However, a contrasting trend emerged in the simulations of SIR model with environmental noise, where variance achieved an AUC of 0.82, exceeding the predictive accuracy of lag-1 AC (Figure \ref{AUC-bar}). 

The PODATR model demonstrated strong performance in certain scenarios, achieving an AUC exceeding 0.95 in the simulations of the SIR models with white noise and environmental noise, surpassing both variance and lag-1 AC (Figure \ref{AUC-bar}). However, its performance decreased in SEIR model, where it achieved an AUC of approximately 0.80 (Figure \ref{AUC-bar}). The model outperformed only variance in the demographic noise model, achieving an AUC of 0.73 (Figure \ref{AUC-bar}). 

The AUC for the SIDATR-500 model consistently reached around one across all test model simulations, illustrating its exceptional predictive capability and robustness. This performance surpassed not only that of the PODATR model but also measures of variance and lag-1 AC, indicating its superiority in capturing the underlying dynamics of the systems under study (Figure \ref{AUC-bar}). On the other hand, the SIDATR-100 model outperformed the variance and PODATR model in the SIR model with demographic noise (Figure \ref{AUC-bar}). Additionally, it outperformed both variance and lag-1 AC in the SIR model with environmental noise (Figure \ref{AUC-bar}).  

\begin{figure}[!tbh]
    \centering
    \includegraphics[width=1.0\linewidth]{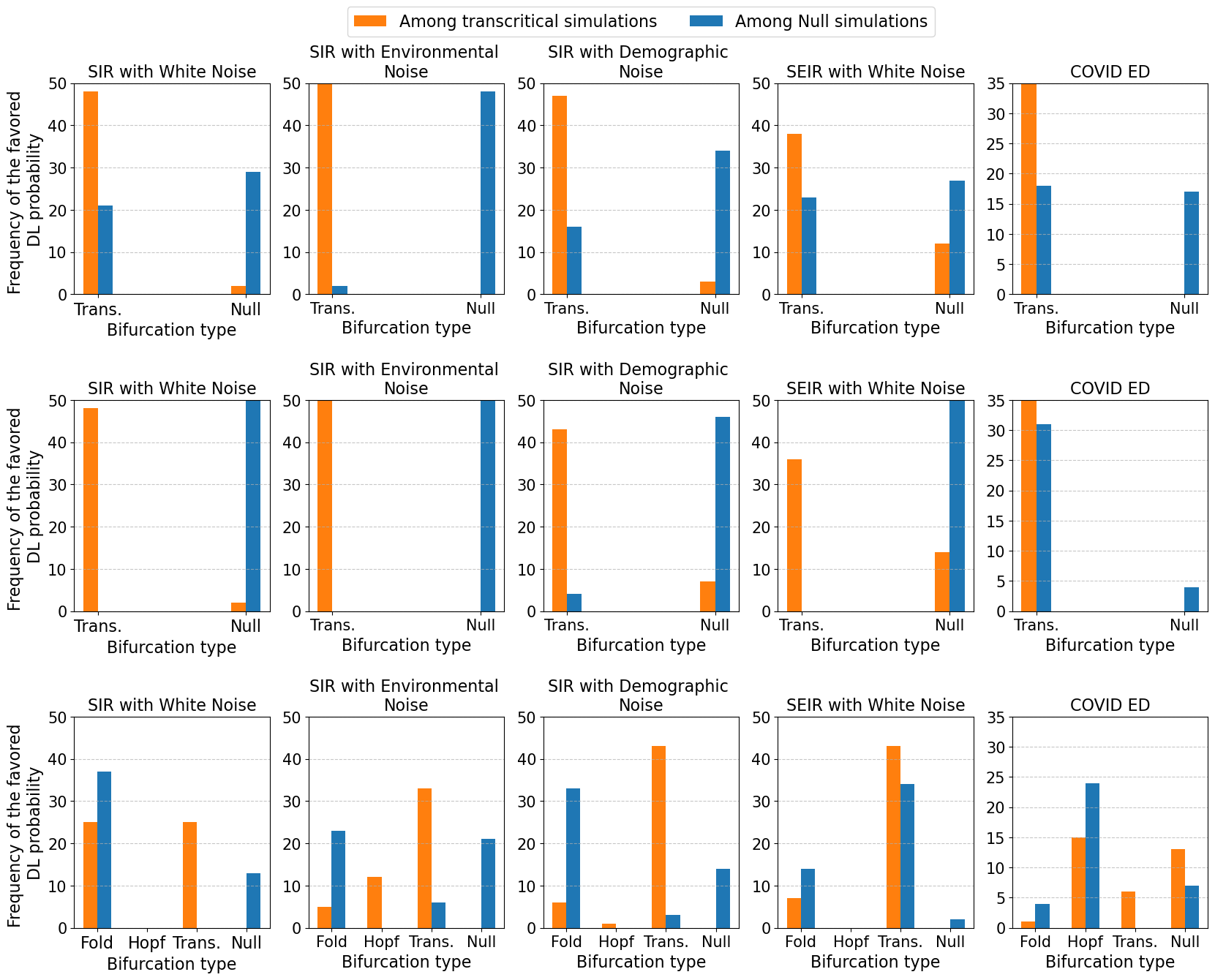}
    \put(-470,330){\textbf{(a)}}
    \put(-470,210){\textbf{(b)}}
    \put(-470, 90){\textbf{(c)}}
    \caption{Frequency distribution of the favored probability among transcritical (blue) and null (orange) simulations generated by the (a) SIDATR-100, (b) SIDATR-500, and (b) PODATR model. In each input simulation, SIDATR-100 and SIDATR-500 models assign probabilities for transcritical and null bifurcations, while PODATR assigns probabilities for fold, Hopf, transcritical, and null bifurcations. Based on the last five favored probabilities in each simulation, a total of 50 frequencies from the transcritical simulations and another 50 frequencies from the null simulations were extracted from each mathematical model. Additionally, for the COVID-19 data, a total of 35 frequencies of each type were extracted.}
    \label{freq-last-20per}
\end{figure}

In transcritical simulations of the SIR model with white noise, both the SIDATR-100 and SIDATR-500 models successfully identified 48 out of 50 frequencies (Figure \ref{freq-last-20per}). However, during null simulations, the SIDATR-100 model detected only 29 frequencies, while the SIDATR-500 model captured all frequencies (Figure \ref{freq-last-20per}). 

In both types of simulations of SIR model with environmental noise, the SIDATR-500 model successfully identified all frequencies, whereas the SIDATR-100 model identified all frequencies in transcritical simulations and 48 out of 50 frequencies in null simulations (Figure \ref{freq-last-20per}). 

Interestingly, in transcritical simulations of both the SIR model with demographic noise and SEIR models, the SIDATR-100 models identified slightly more frequencies compared to the SIDATR-500 models, whereas in null simulations, the SIDATR-500 model exhibited better frequency detection than the SIDATR-100 model (Figure \ref{freq-last-20per}). 

The PODATR model consistently identified impending transitions in transcritical simulations across all test models, demonstrating its effectiveness in recognizing critical changes in the system. However, it encountered challenges in accurately classifying most of the null simulations within the test models. Despite its success in identifying impending transitions in transcritical simulations, the PODATR model occasionally misclassified bifurcation types among frequencies (Figure \ref{freq-last-20per}).

Overall, the SIDATR-500 model consistently exhibited superior frequency performance in the last five predictions across the simulated test models compared to both the PODATR and SIDATR-100 models (Figure \ref{freq-last-20per}). However, in most simulations, particularly during shorter time spans and farther from the bifurcation point, the SIDATR-500 model displayed variations in probability assignments (Supplementary Figures \ref{prob_white_noise_trans}, \ref{prob_env_noise_trans}, \ref{prob_dem_noise_trans}, and \ref{prob_SEIR_trans}). Nonetheless, as the simulations progressed toward the bifurcation point, the model's predictions significantly improved, showcasing precise probability assignments for the correct bifurcation type.

\subsection{Performance on empirical data}
We employed seven transcritical and null time series from the COVID-19 dataset of Edmonton as inputs for the generic EWIs and the trained DL models and investigated the EWSs of the outbreak. The AUC values for variance, lag-1 AC, the PODATR model, and the SIDATR-500 model were all found below 0.50 (Figure \ref{AUC-bar}). Notably, the SIDATR-500 model showed an improvement in AUC compared to other indicators, yet overall performance remained below random chance. Intriguingly, the SIDATR-100 model demonstrated superior performance within the COVID-19 dataset, achieving an AUC of 0.71 and outperforming other indicators (Figure \ref{AUC-bar}).

Each set of transcritical and null simulations yielded 35 frequencies derived from the seven time series. The PODATR model successfully predicted transitions in over half of the frequencies within transcritical simulations (Figure \ref{freq-last-20per}). However, in null simulations, most frequencies were inaccurately identified (Figure \ref{freq-last-20per}). Among the accurately identified frequencies in transcritical simulations, the majority corresponded to the Hopf bifurcation type (Figure \ref{freq-last-20per}). In contrast, the SIDATR-500 model successfully predicted all frequencies within transcritical simulations but struggled to accurately identify most frequencies in null simulations (Figure \ref{freq-last-20per}). The SIDATR-100 model exhibited a superior frequency distribution compared to the other models, accurately capturing all frequencies within transcritical simulations and approximately half of the frequencies within null simulations (Figure \ref{freq-last-20per}). These findings suggest that the DL algorithm, trained by the noise-induced model, provides enhanced EWSs for impending transitions of a disease outbreak. 

\section{Discussion}

We investigated EWSs in the context of noise-induced disease-spreading models, incorporating additive white noise, multiplicative environmental noise, and demographic noise. We analyzed these signals by employing a DL algorithm alongside two standard EWIs: variance and lag-1 AC. To effectively train the DL algorithm, we introduced a training dataset originating from noise-induced SIR models. We tested both the trained DL algorithm and generic EWIs over the noise-induced disease models, scrutinizing their EWSs in scenarios with and without critical transitions. We also assessed a pre-trained DL algorithm from Bury et al. \cite{bury2021deep}, trained on simulated data from randomly generated dynamical systems characterized by general polynomial terms. For a comprehensive evaluation, we compared the area under the ROC curve of all the indicators. In assessing the DL algorithm's performance on real-world disease outbreaks, we employed an empirical COVID-19 cases dataset. Remarkably, the DL algorithm demonstrated significant improvement across all test models and empirical data when trained on noise-induced disease model data, outperforming other EWIs.

The performance of the classifiers, except for SIDATR-100, fell below random chance when evaluated on the empirical COVID-19 dataset from Edmonton. While the SIDATR-500 model demonstrated superior performance in transcritical simulations of empirical data, it performed poorly in null simulations. This inconsistency is associated with the short length of the empirical dataset, particularly in null simulations. The PODATR and SIDATR-500 models were trained on 500 time points, whereas the null simulations of COVID-19 consisted of less than 100 time points. Conversely, the performance notably improved with the SIDATR-100 model, which was trained on a shorter time span of 100 points. Consequently, enhancing the model's performance on empirical data can be achieved by training it on shorter time lengths, as demonstrated by the efficacy of the SIDATR-100 model. 

The initial probabilities assigned by the SIDATR-500 model were found to be inaccurate for a very short time series. However, the model's predictive performance significantly improved as the observed time point neared the bifurcation point. This inconsistency can be explained to the DL model's enhanced ability to provide more accurate probabilities when it is close to the transition point \cite{deb2022machine}, whereas the initial predictions may include data points that are considerably distant from this transition point. This phenomenon aligns with the concept of critical slowing down in a dynamical system. Critical slowing down typically manifests when a system approaches a bifurcation point, leading to the emergence of distinctive patterns in the time series \cite{scheffer2012anticipating, boettiger2013early}. However, when analyzing a lengthy time series that does not sufficiently approach the bifurcation point, these characteristic patterns of critical slowing down may not be readily apparent.

The capability of the SIDATR-500 and SIDATR-100 models to identify bifurcations in a time series beyond those on which they were specifically trained was assessed using an SEIR model with additive white noise and real-world COVID-19 cases. The SEIR model, characterized by a higher dimensionality compared to the training models, serves as a versatile framework for describing the dynamics of various infectious diseases \cite{brauer2008mathematical}, and the COVID-19 pandemic stands as a recent and significant public health challenge. The performance of generating EWSs of these DL models can be ascribed to the diverse training data deliberately generated with a wide array of random parameters, encompassing all conceivable scenarios that may manifest in a disease outbreak. Additionally, by incorporating different types of noise with random intensity, our training data replicated the varying levels of fluctuation, mirroring the uncertainty inherent in real-world outbreaks \cite{huber2000effects}. Notably, our training data accounted for the linear increase in disease transmission rate, reflecting the common trend of initially low infection levels that intensify over time in infectious diseases \cite{thurner2020network}. Moreover, we accommodated cases where the rate increased but did not reach the threshold for a full-blown outbreak. By incorporating all these potential phenomena in the training data, the DL model gained the necessary adaptability to provide EWSs in models that closely resemble real-world scenarios.

One of the hypotheses examined in \cite{bury2021deep} was whether the DL algorithm could provide more effective EWSs compared to generic EWIs. Within the context of our study, the performance of the PODATR model both supported and contradicted this hypothesis in different test models. The training data for the PODATR model was derived from a two-dimensional random dynamical system featuring a general polynomial up to the third order and incorporating additive white noise in the simulations. The results revealed that the performance of the PODATR model was less effective than lag-1 AC, SIDATR-100, and SIDATR-500 models for the simulations of SIR with demographic noise. Notably, infectious diseases often involve a range of noise sources in their dynamics \cite{sabbar2022influence}, and these noises can significantly impact the robustness of early warning systems \cite{qin2018early}. Beside these noises, mathematical models are generally built on a set of underlying assumptions, and the dynamics of population compartments are contingent on these assumptions \cite{brauer2008mathematical}. As a result, certain models may exhibit distinct dynamical behavior, potentially leading to challenges for the trained model in accurately capturing trends in those cases. As evidenced by the results, the PODATR model exhibited the lowest accuracy compared to some other indicators in the SEIR model, even when considering only additive white noise. 

The performance of Lag-1 AC and variance exhibited variations in the SIR model with environmental and demographic noise. Specifically, variance demonstrated superior performance in one model, while Lag-1 AC outperformed it in the other. This variability in performance can be attributed to the differing levels of noise intensities present in the simulations and different types of noises in the system \cite{liu2015identifying}. The simulations used for performance assessment were generated by random noise intensity, and it is worth noting that the generic EWIs are sensitive to noise intensity \cite{proverbio2023systematic} as well as the type of noise \cite{o2018stochasticity}, underscoring the need to explore alternative statistical measures for noise-induced models in future research.

The SIR model is a generalized framework for characterizing a wide range of infectious diseases \cite{brauer2008mathematical}, with additive white noise, multiplicative environmental noise, and demographic noise representing common types of noise encountered in disease outbreaks \cite{allen2010introduction}. Consequently, the noise-induced SIR models employed in this study serve as fundamental representations capable of capturing the dynamics of various infectious disease outbreaks. By amalgamating data from a hundred thousand simulations across all three models, we introduced a substantial level of variability into the training dataset. This diversity equips the DL model with the ability to adapt to a spectrum of scenarios. Since DL models typically excel when applied to data that closely resembles their training dataset \cite{deb2022machine}, the SIDATR-100 and SIDATR-500 models, having been trained on data closely aligned with these noise-induced models, possess the adaptability to exhibit heightened sensitivity and specificity in providing EWSs for emerging and re-emerging infectious disease outbreaks that exhibit similarities with these noise-induced models.

Our study has several limitations. First, despite having a maximum simulation length of 1500 time points, the DL algorithm was exclusively trained on a subset of observations consisting of 500 and 100 time points. This limited exposure to both shorter and longer time points, potentially restricting its ability to generalize across a broader range of temporal dynamics. Second, the trained DL model's sensitivity to data peaks could lead to misclassification when applied to null simulations, as these simulations do not exhibit peaks over time. Third, our training process primarily focused on linear variations in the disease transmission rate, overlooking other forms of rate variations and potential fluctuations in other parameters within the SIR model, which may be crucial in real-world scenarios \cite{long2021identification}. Fourth, our training data pertained to three specific types of noise-induced SIR models, which might limit the model's suitability for capturing transitions in models with different behaviors. Although we aimed to closely simulate real-world scenarios, it is essential to note that our model was not specifically trained on empirical data. Future directions should explore training on empirical datasets to assess models' performance in real-world scenarios. Fifth, we explored only one DL algorithm and did not investigate the capabilities of other DL architectures. This will be a focus of future research. Sixth, it is important to acknowledge that none of the models can predict the exact timing of bifurcations, highlighting the need for further investigation in this area. 

This research has a significant impact on the generation of reliable EWSs for impending transitions in infectious disease outbreaks. While acknowledging the model's limitations and offering suggestions, it opens the door for future research aimed at developing even more effective early warning systems.

\section*{Data availability}
Training data are available in a Zenodo repository (\url{https://doi.org/10.5281/zenodo.10841970})\cite{chakraborty_2024_10841970}.
Testing codes and other simulations have been deposited in GitHub (\url{https://github.com/AmitKChakraborty/EWSofInfectiousDiseases}). The COVID-19 data are available at the city of Edmonton's open data portal \cite{COVID-19Edmonton}.

\subsection*{Funding}
This project (Amit K. Chakraborty, Shan Gao, Reza Miry) was primarily supported by One Health Modelling Network for Emerging Infections (OMNI), Amii (Alberta Machine Intelligence Institute) matching fund, and the Department of Mathematical and Statistical Sciences (IUSEP funding) at the University of Alberta. Hao Wang's research was partially supported by the Natural Sciences and Engineering Research Council of Canada (Individual Discovery Grant RGPIN-2020-03911 and Discovery Accelerator Supplement Award RGPAS-2020-00090) and the Canada Research Chairs program (Tier 1 Canada Research Chair Award). Mark A. Lewis gratefully acknowledges support from an NSERC Discovery Grant and the Gilbert and Betty Kennedy Chair in Mathematical Biology. Pouria Ramazi also acknowledges the support from an NSERC Discovery Grant.

\section*{Acknowledgement}
We would like to thank Pijush Panday for the initial discussion and Tianyu Guan for providing valuable feedback. 

\section*{Declaration of AI use}
We have not used AI-assisted technologies in creating this article.

\section*{Competing interests} We declare we have no competing interests.

\bibliographystyle{unsrt}
\bibliography{2references-ews}

\end{document}



\maketitle

\section*{Supplementary figures}
\setcounter{table}{0}  
\setcounter{figure}{0}

\renewcommand{\thetable}{S\arabic{table}}
\renewcommand{\thefigure}{S\arabic{figure}}

In this document, we present the supporting figures for the manuscript titled "An early warning indicator trained on stochastic disease-spreading models with different noise levels."

\begin{figure}[!tbh]
    \centering
    \includegraphics[width=1.0\linewidth]{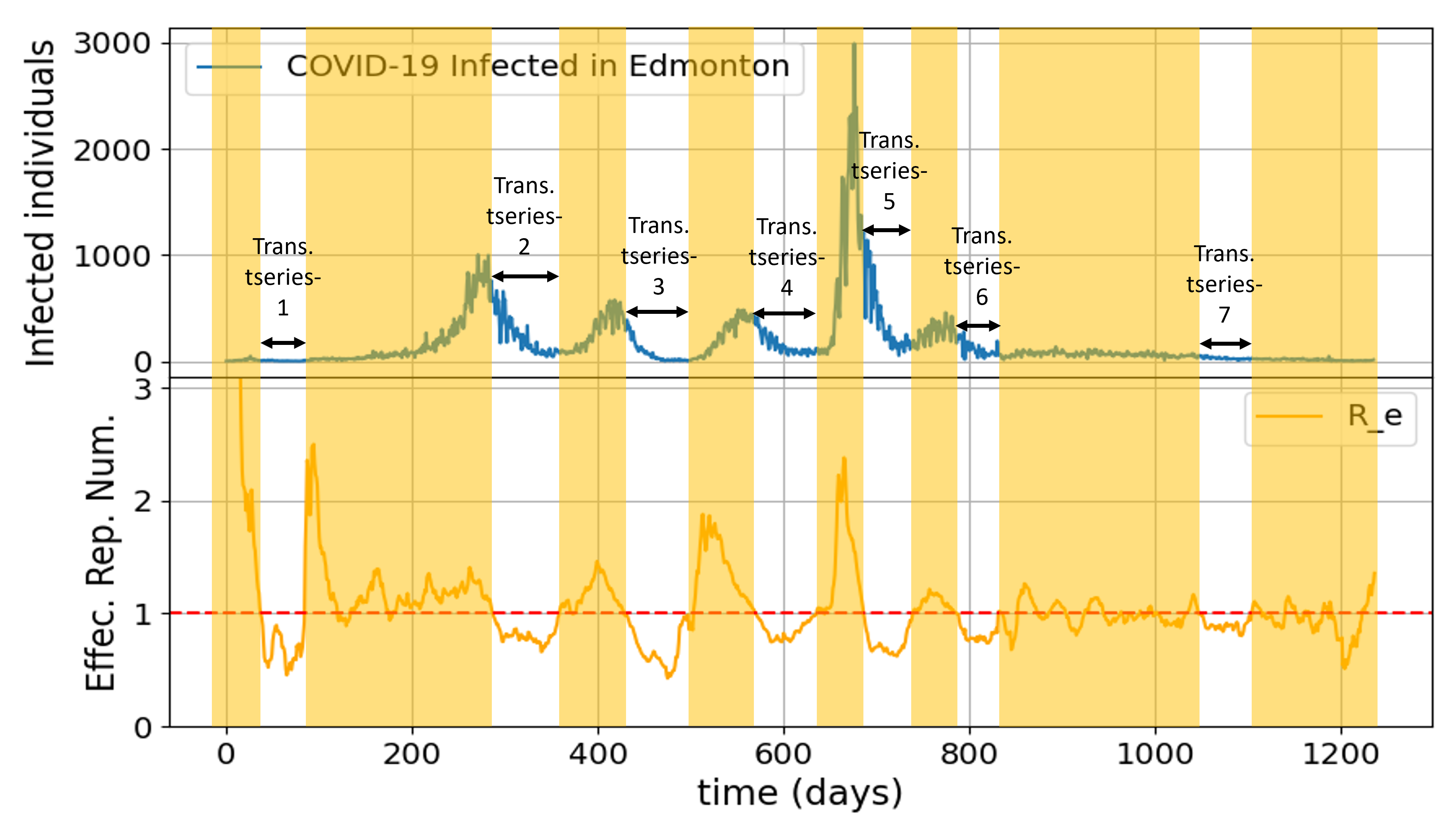}
    \caption{Empirical data of COVID-19 in Edmonton and the calculated effective reproduction number using \texttt{EpiEstim} package. Based on $R_e < 1$, we extracted daily cases until $R_e = 1$. Each sub-series is labeled as transcritical, representing the cases just before the bifurcation point, i.e., $R_e = 1$.}
    \label{empirical_covid_Re}
\end{figure}

\begin{figure}[ht!]
\centering
 \begin{subfigure}[t]{0.19\textwidth}
    \includegraphics[width=\textwidth]{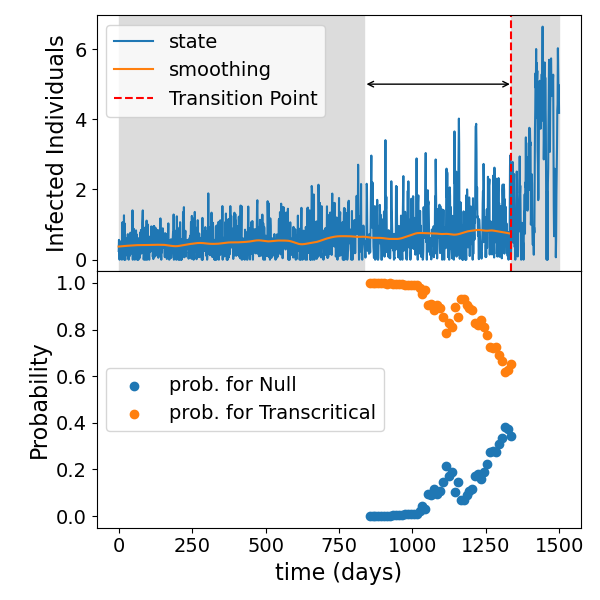}
 \end{subfigure}
 \hfill 
 \begin{subfigure}[t]{0.19\textwidth}
    \includegraphics[width=\textwidth]{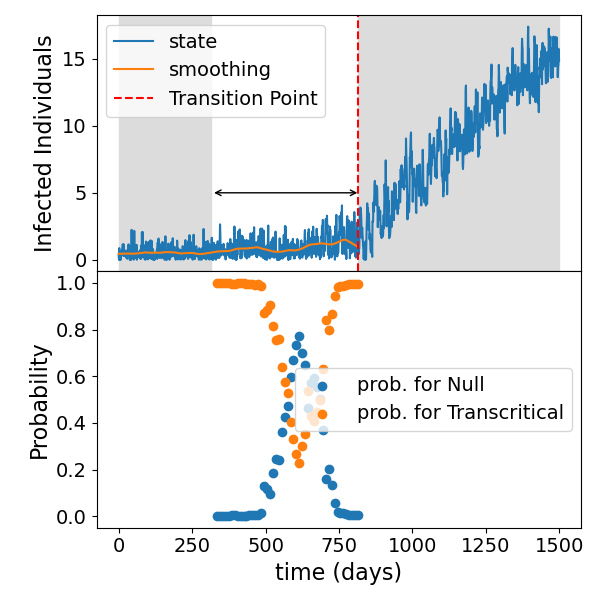}
 \end{subfigure}
  \hfill 
 \begin{subfigure}{0.19\textwidth}
     \includegraphics[width=\textwidth]{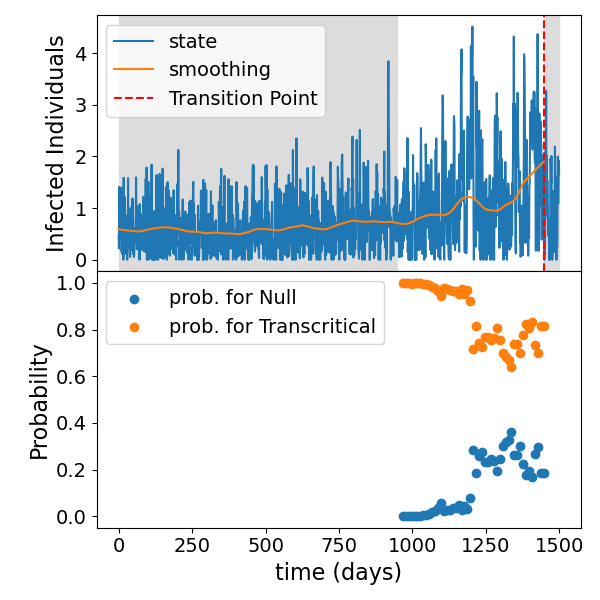}
 \end{subfigure}
  \hfill 
  \begin{subfigure}[t]{0.19\textwidth}
    \includegraphics[width=\textwidth]{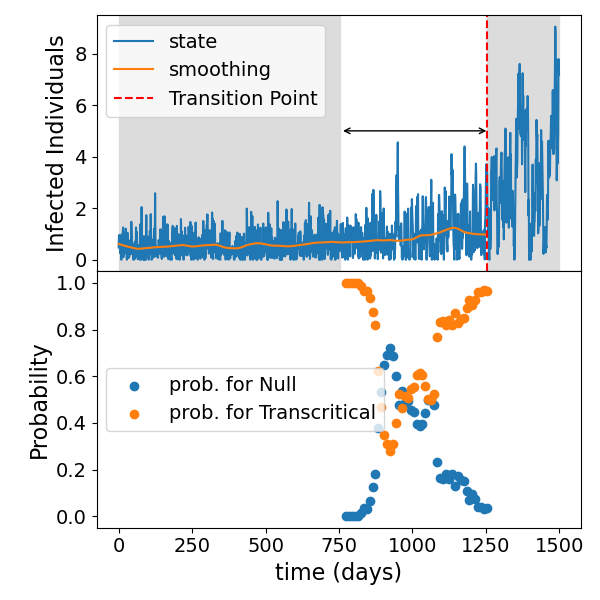}
 \end{subfigure}
  \hfill 
 \begin{subfigure}{0.19\textwidth}
     \includegraphics[width=\textwidth]{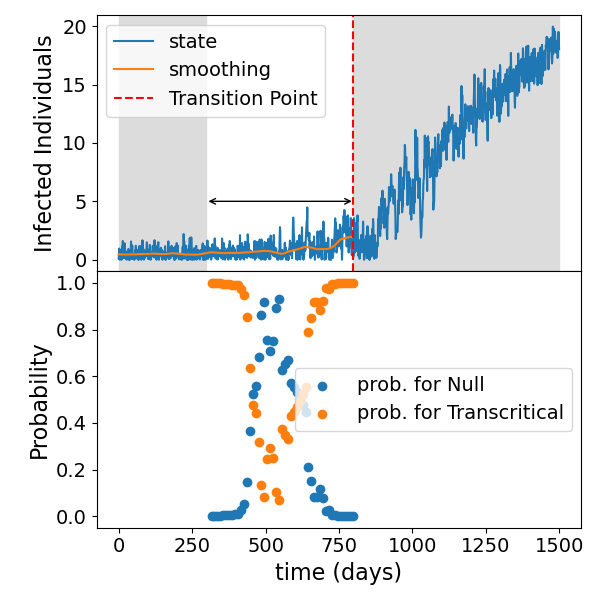}
 \end{subfigure} 
\medskip
  \begin{subfigure}[t]{0.19\textwidth}
    \includegraphics[width=\textwidth]{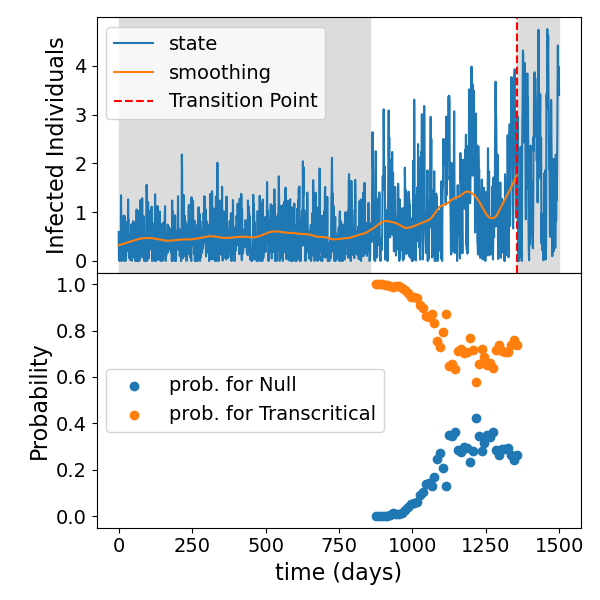}
 \end{subfigure}
  \hfill 
 \begin{subfigure}[t]{0.19\textwidth}
    \includegraphics[width=\textwidth]{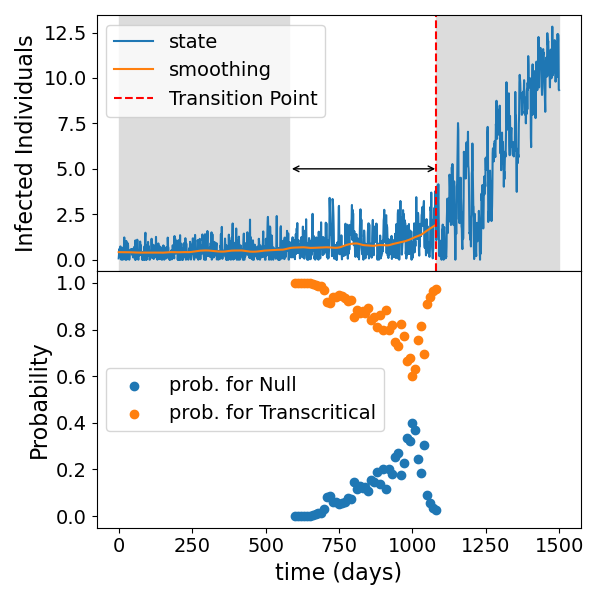}
 \end{subfigure}
  \hfill 
 \begin{subfigure}{0.19\textwidth}
     \includegraphics[width=\textwidth]{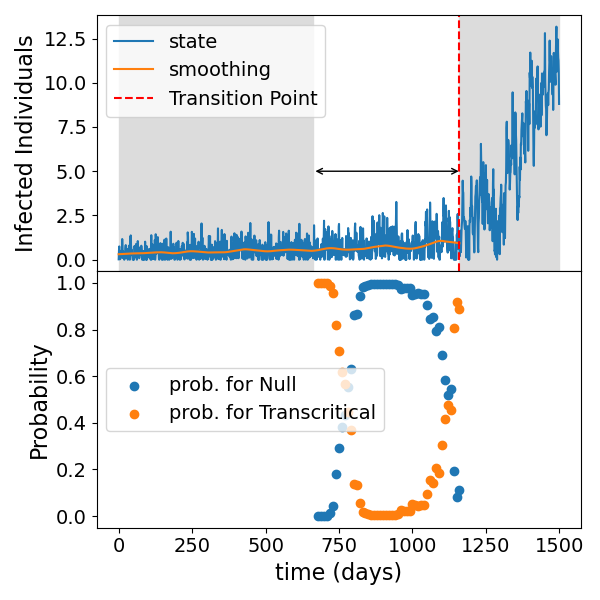}
 \end{subfigure}
  \hfill 
  \begin{subfigure}[t]{0.19\textwidth}
    \includegraphics[width=\textwidth]{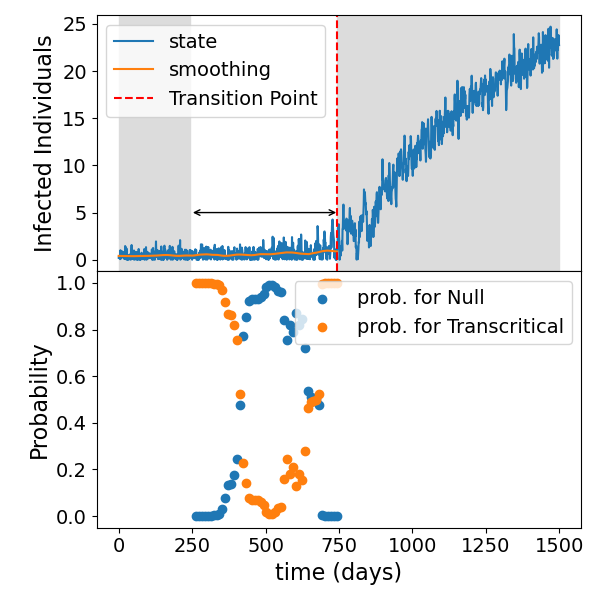}
 \end{subfigure}
  \hfill 
 \begin{subfigure}{0.19\textwidth}
    \includegraphics[width=\textwidth]{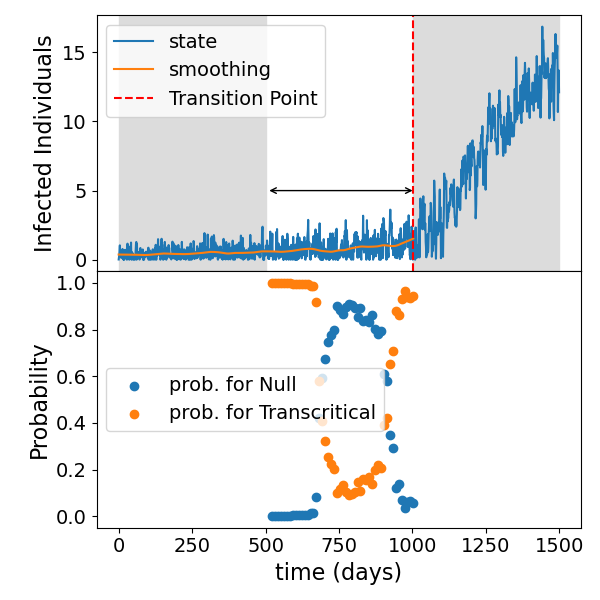}
 \end{subfigure}
\caption{Probabilities for a transition assigned by the SIDATR-500 DL model based on the subset of observations of transcritical simulations of the SIR model with additive white noise.}
\label{prob_white_noise_trans}
\end{figure}

\begin{figure}
\centering
 \begin{subfigure}[t]{0.19\textwidth}
    \includegraphics[width=\textwidth]{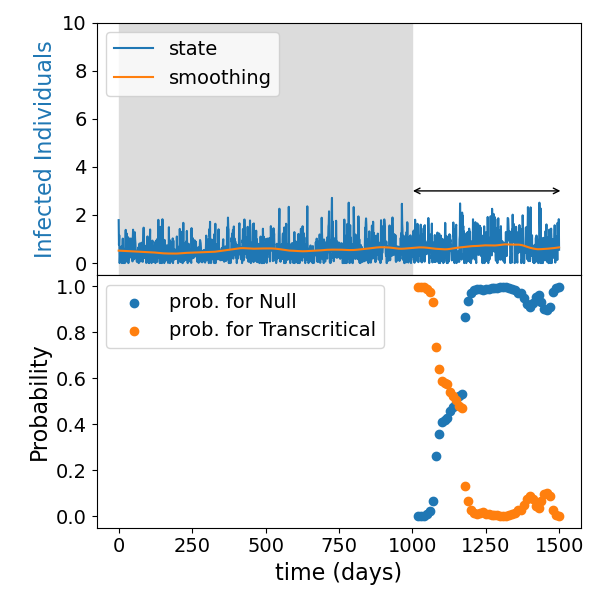}
 \end{subfigure}
 \hfill 
 \begin{subfigure}[t]{0.19\textwidth}
    \includegraphics[width=\textwidth]{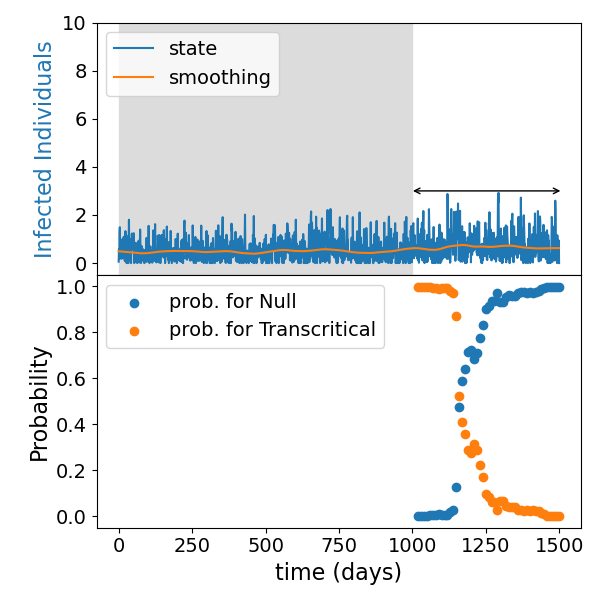}
 \end{subfigure}
  \hfill 
 \begin{subfigure}{0.19\textwidth}
     \includegraphics[width=\textwidth]{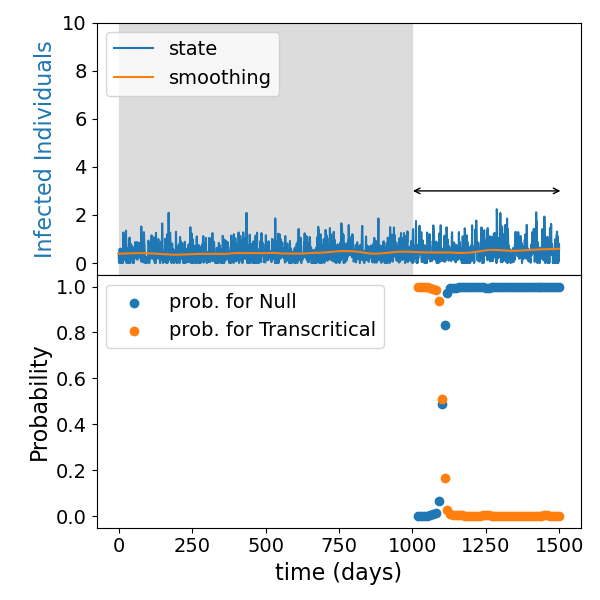}
 \end{subfigure}
  \hfill 
  \begin{subfigure}[t]{0.19\textwidth}
    \includegraphics[width=\textwidth]{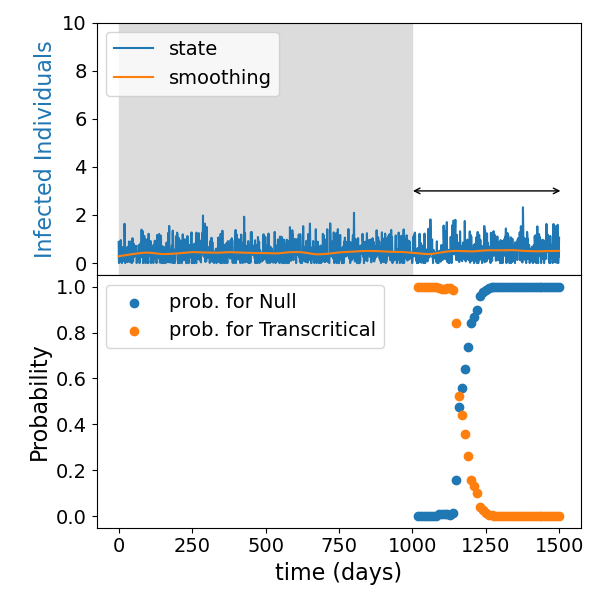}
 \end{subfigure}
  \hfill 
 \begin{subfigure}{0.19\textwidth}
     \includegraphics[width=\textwidth]{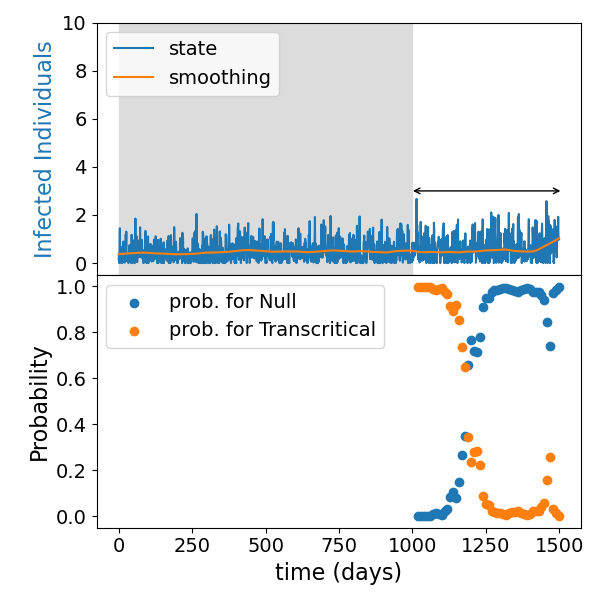}
 \end{subfigure} 
\medskip
  \begin{subfigure}[t]{0.19\textwidth}
    \includegraphics[width=\textwidth]{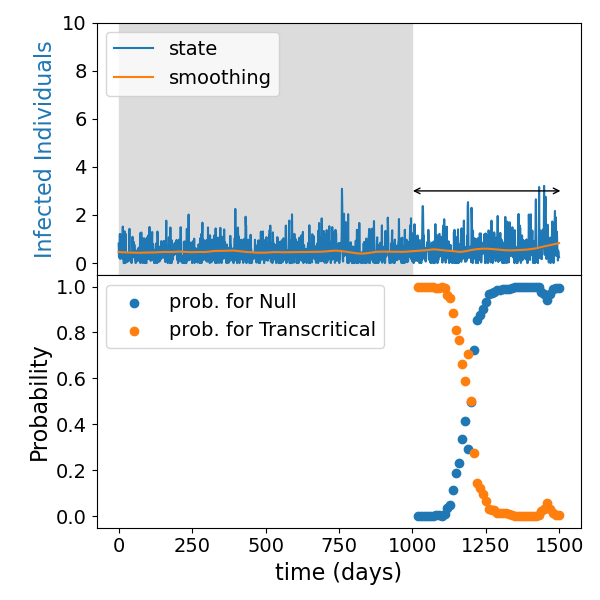}
 \end{subfigure}
  \hfill
 \begin{subfigure}[t]{0.19\textwidth}
    \includegraphics[width=\textwidth]{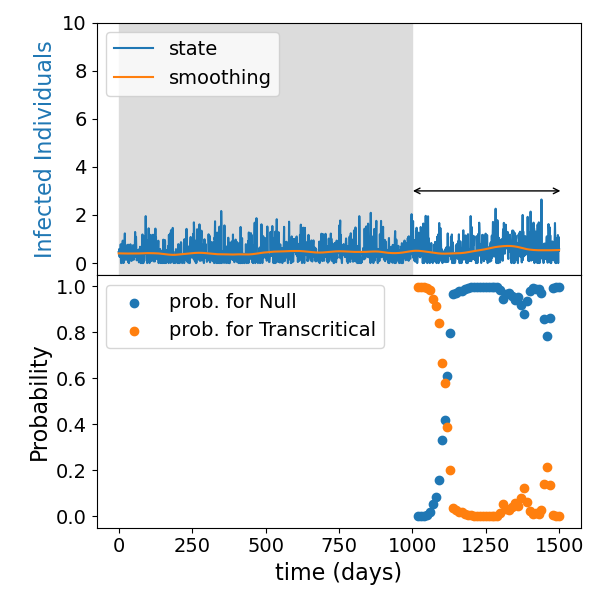}
 \end{subfigure}
  \hfill 
 \begin{subfigure}{0.19\textwidth}
     \includegraphics[width=\textwidth]{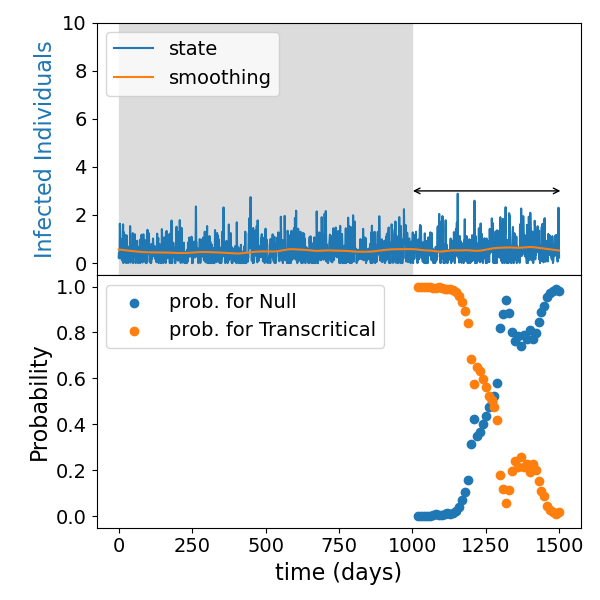}
 \end{subfigure}
  \hfill 
  \begin{subfigure}[t]{0.19\textwidth}
    \includegraphics[width=\textwidth]{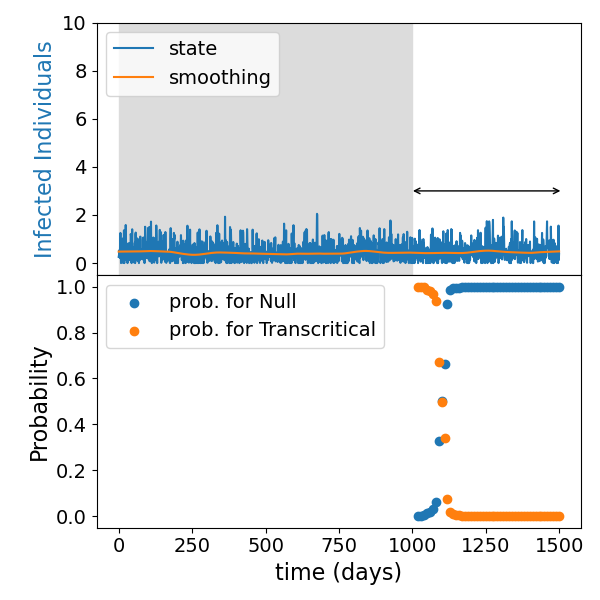}
 \end{subfigure}
  \hfill 
 \begin{subfigure}{0.19\textwidth}
    \includegraphics[width=\textwidth]{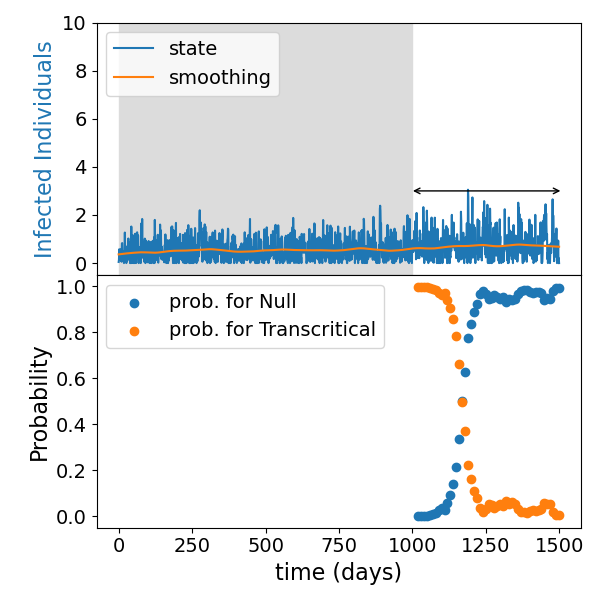}
 \end{subfigure}
\caption{Probabilities for a transition assigned by the SIDATR-500 DL model based on the subset of observations of null simulations of the SIR model with additive white noise.}
\label{prob_white_noise_null}
\end{figure}

\begin{figure}
\centering
 \begin{subfigure}[t]{0.19\textwidth}
    \includegraphics[width=\textwidth]{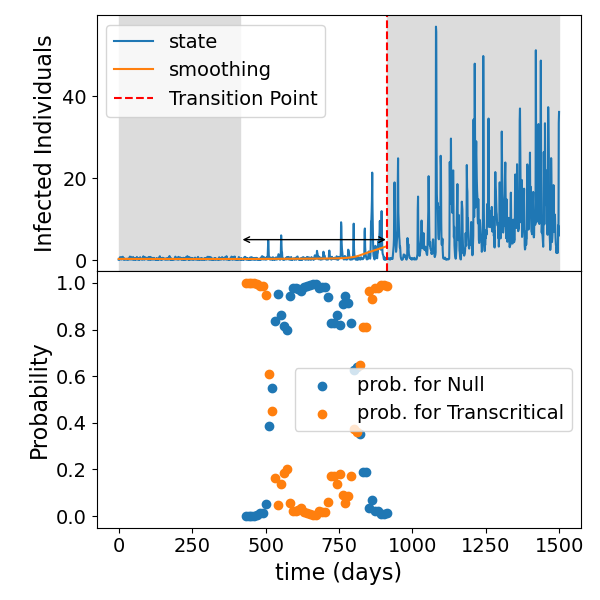}
 \end{subfigure}
 \hfill 
 \begin{subfigure}[t]{0.19\textwidth}
    \includegraphics[width=\textwidth]{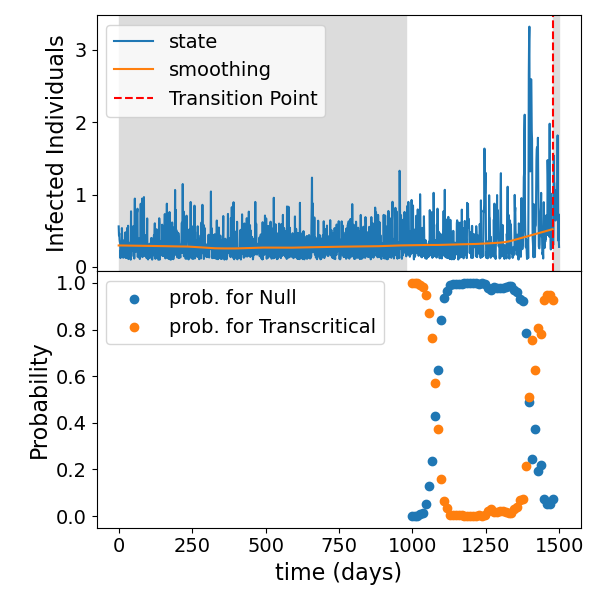}
 \end{subfigure}
  \hfill 
 \begin{subfigure}{0.19\textwidth}
     \includegraphics[width=\textwidth]{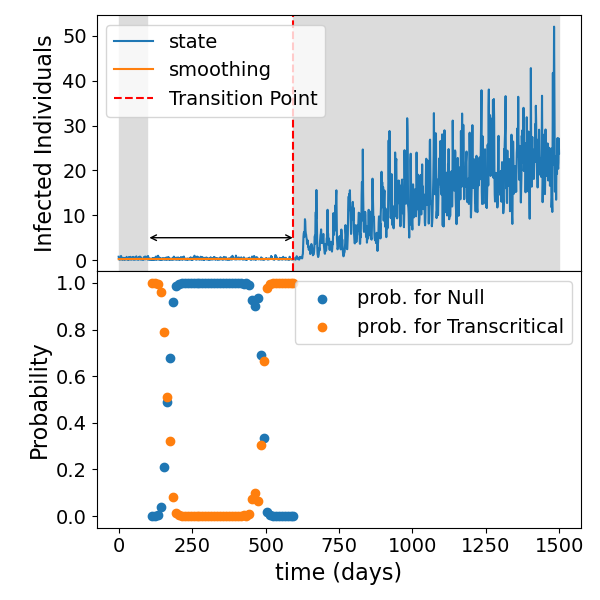}
 \end{subfigure}
  \hfill 
  \begin{subfigure}[t]{0.19\textwidth}
    \includegraphics[width=\textwidth]{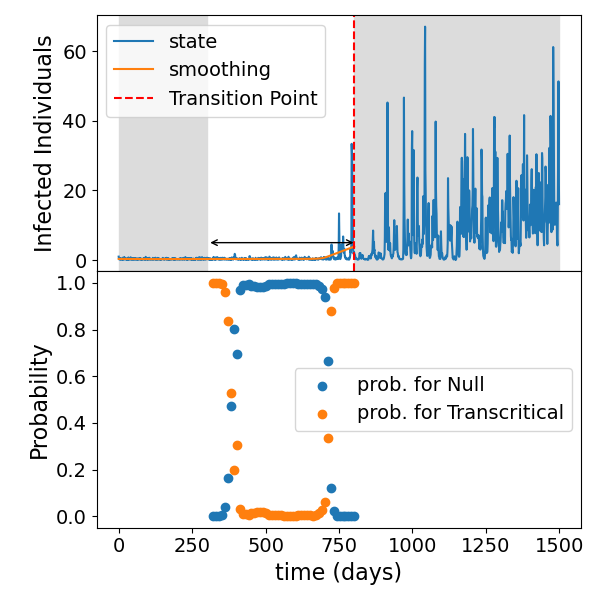}
 \end{subfigure}
  \hfill 
 \begin{subfigure}{0.19\textwidth}
     \includegraphics[width=\textwidth]{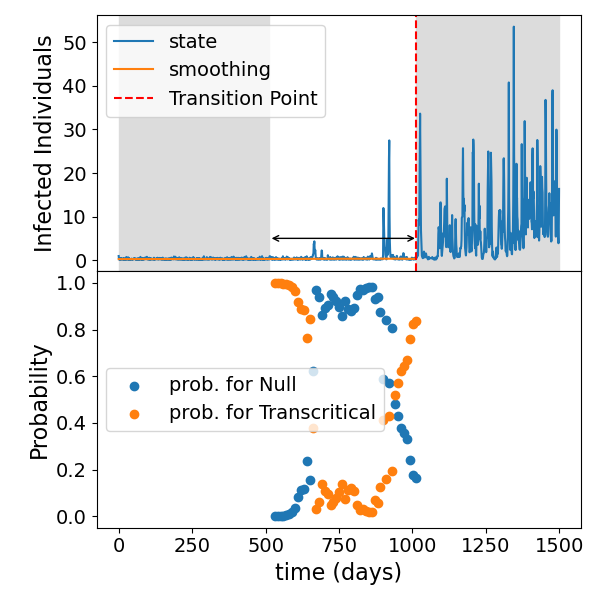}
 \end{subfigure} 
\medskip
  \begin{subfigure}[t]{0.19\textwidth}
    \includegraphics[width=\textwidth]{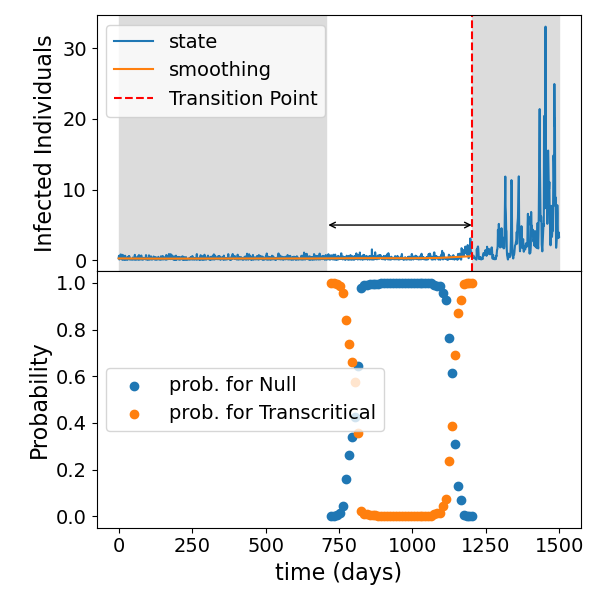}
 \end{subfigure}
  \hfill 
 \begin{subfigure}[t]{0.19\textwidth}
    \includegraphics[width=\textwidth]{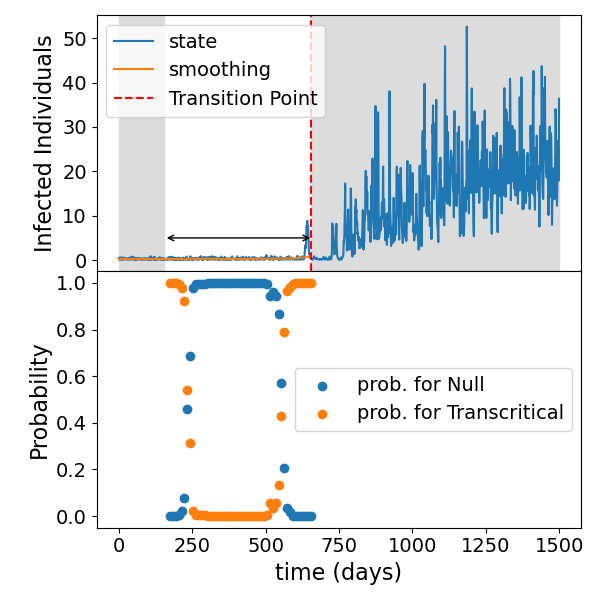}
 \end{subfigure}
  \hfill 
 \begin{subfigure}{0.19\textwidth}
     \includegraphics[width=\textwidth]{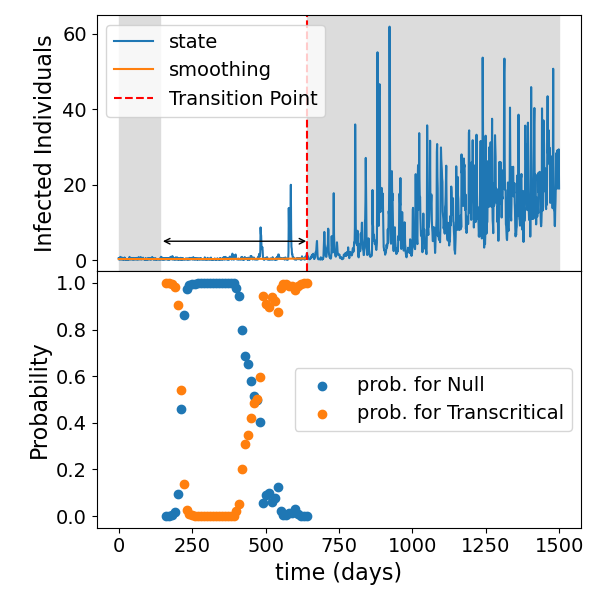}
 \end{subfigure}
  \hfill 
  \begin{subfigure}[t]{0.19\textwidth}
    \includegraphics[width=\textwidth]{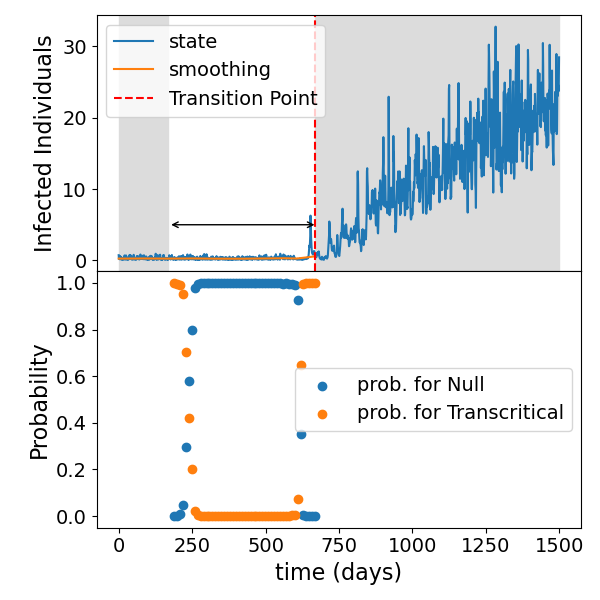}
 \end{subfigure}
  \hfill 
 \begin{subfigure}{0.19\textwidth}
    \includegraphics[width=\textwidth]{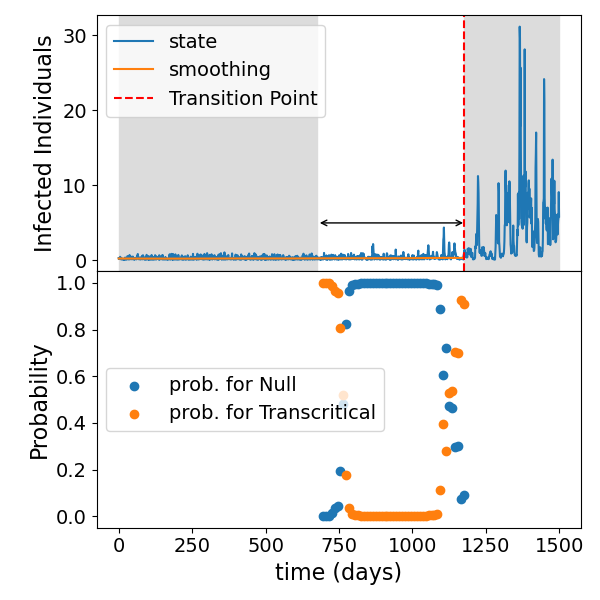}
 \end{subfigure}
\caption{Probabilities for a transition assigned by the SIDATR-500 DL model based on the subset of observations of transcritical simulations of the SIR model with multiplicative environmental noise.}
\label{prob_env_noise_trans}
\end{figure}

\begin{figure}
\centering
 \begin{subfigure}[t]{0.19\textwidth}
    \includegraphics[width=\textwidth]{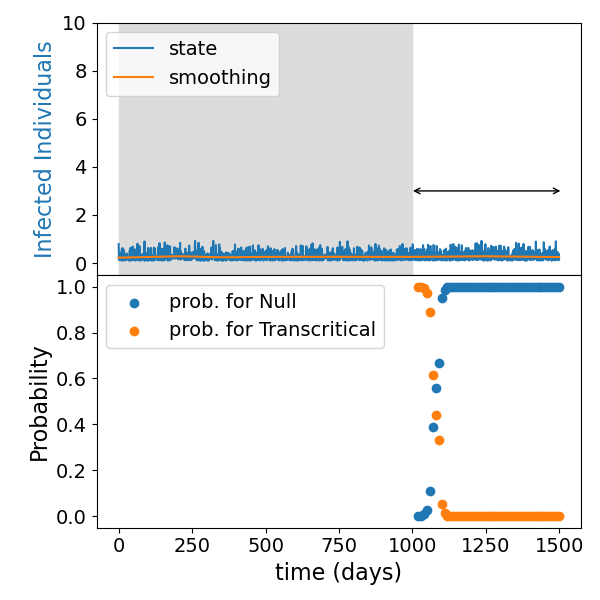}
 \end{subfigure}
 \hfill 
 \begin{subfigure}[t]{0.19\textwidth}
    \includegraphics[width=\textwidth]{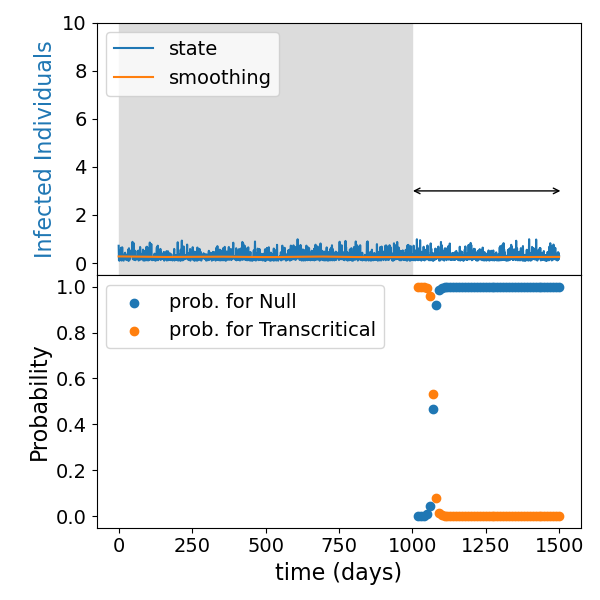}
 \end{subfigure}
  \hfill 
 \begin{subfigure}{0.19\textwidth}
     \includegraphics[width=\textwidth]{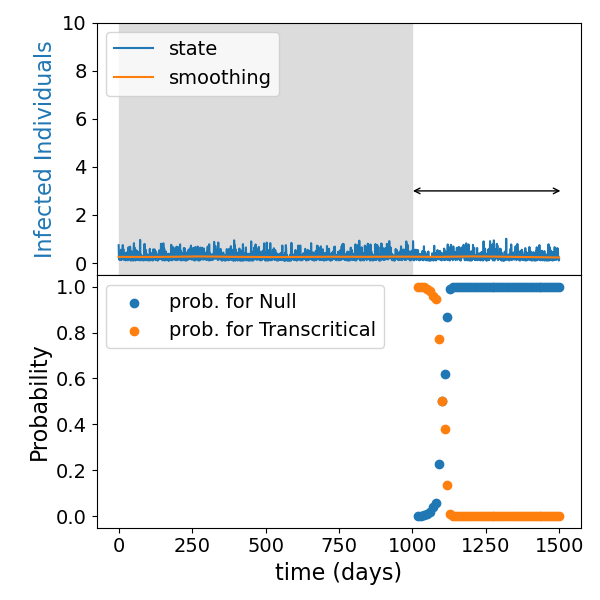}
 \end{subfigure}
  \hfill 
  \begin{subfigure}[t]{0.19\textwidth}
    \includegraphics[width=\textwidth]{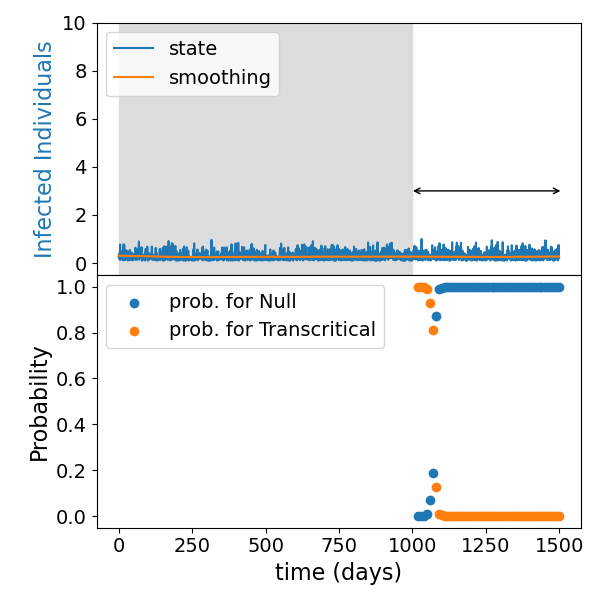}
 \end{subfigure}
  \hfill 
 \begin{subfigure}{0.19\textwidth}
     \includegraphics[width=\textwidth]{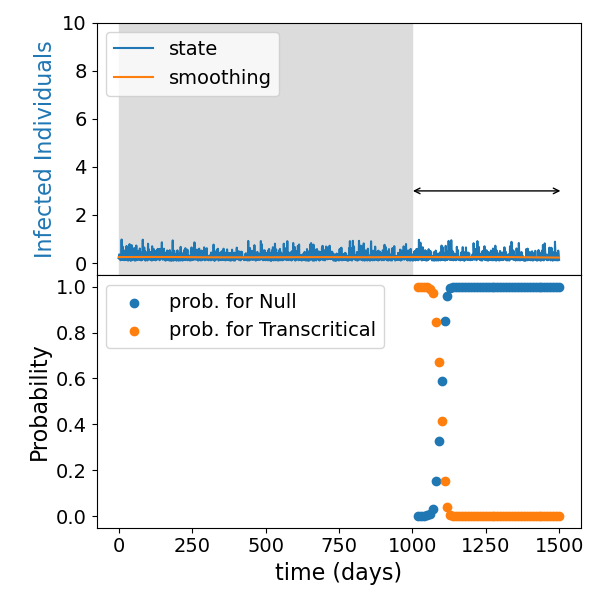}
 \end{subfigure} 
\medskip
  \begin{subfigure}[t]{0.19\textwidth}
    \includegraphics[width=\textwidth]{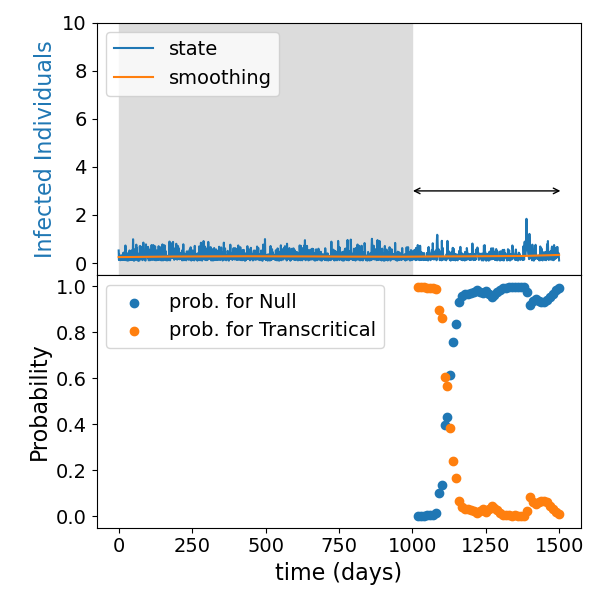}
 \end{subfigure}
  \hfill 
 \begin{subfigure}[t]{0.19\textwidth}
    \includegraphics[width=\textwidth]{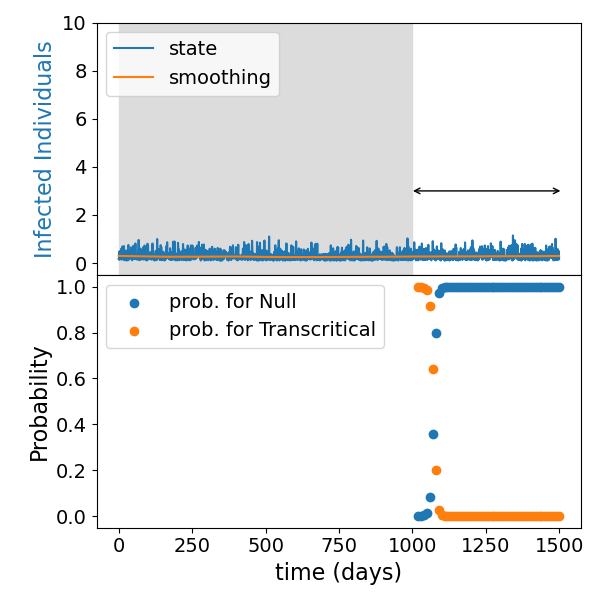}
 \end{subfigure}
  \hfill 
 \begin{subfigure}{0.19\textwidth}
     \includegraphics[width=\textwidth]{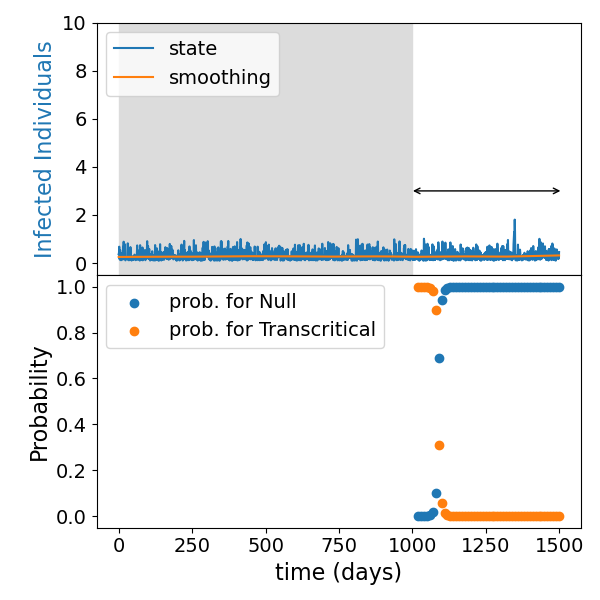}
 \end{subfigure}
  \hfill 
  \begin{subfigure}[t]{0.19\textwidth}
    \includegraphics[width=\textwidth]{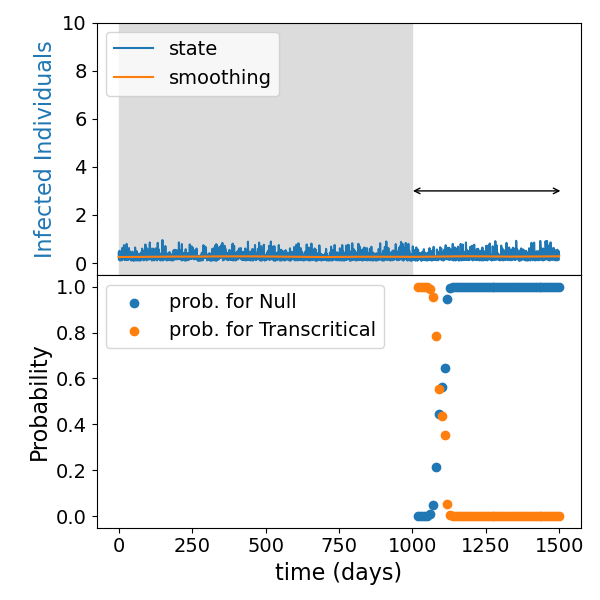}
 \end{subfigure}
  \hfill 
 \begin{subfigure}{0.19\textwidth}
    \includegraphics[width=\textwidth]{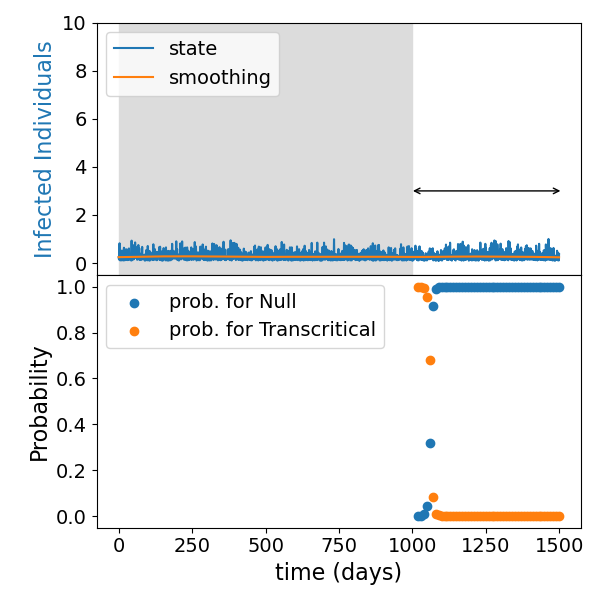}
 \end{subfigure}
\caption{Probabilities for a transition assigned by the SIDATR-500 DL model based on the subset of observations of null simulations of the SIR model with multiplicative environmental noise.}
\label{prob_env_noise_null}
\end{figure}

\begin{figure}
\centering
 \begin{subfigure}[t]{0.19\textwidth}
    \includegraphics[width=\textwidth]{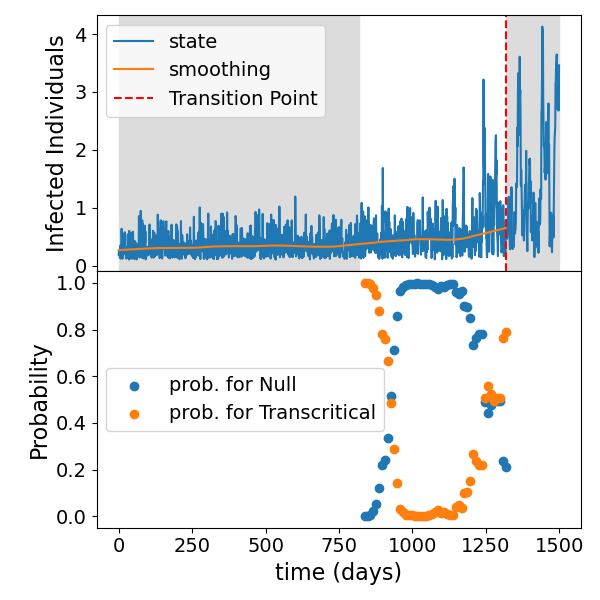}
 \end{subfigure}
 \hfill 
 \begin{subfigure}[t]{0.19\textwidth}
    \includegraphics[width=\textwidth]{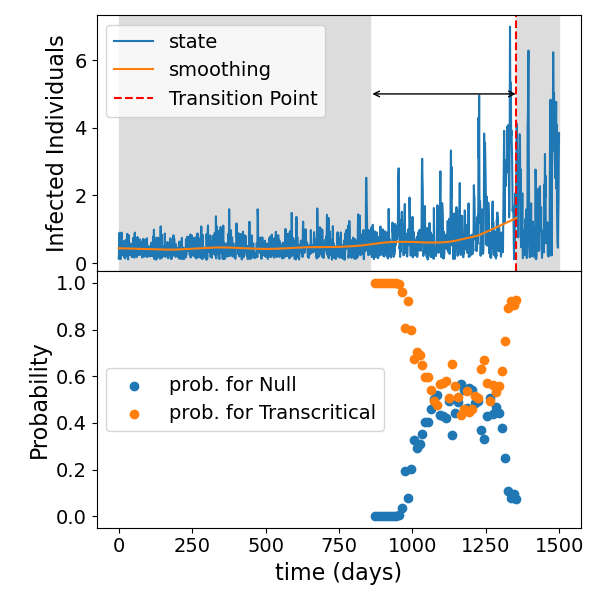}
 \end{subfigure}
  \hfill 
 \begin{subfigure}{0.19\textwidth}
     \includegraphics[width=\textwidth]{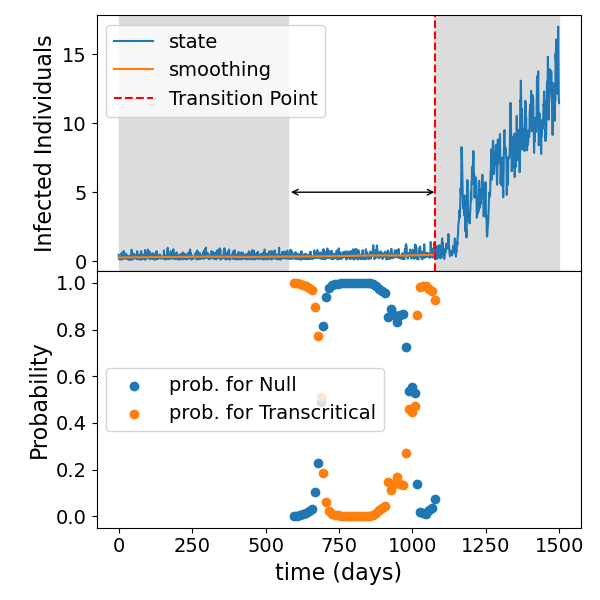}
 \end{subfigure}
  \hfill 
  \begin{subfigure}[t]{0.19\textwidth}
    \includegraphics[width=\textwidth]{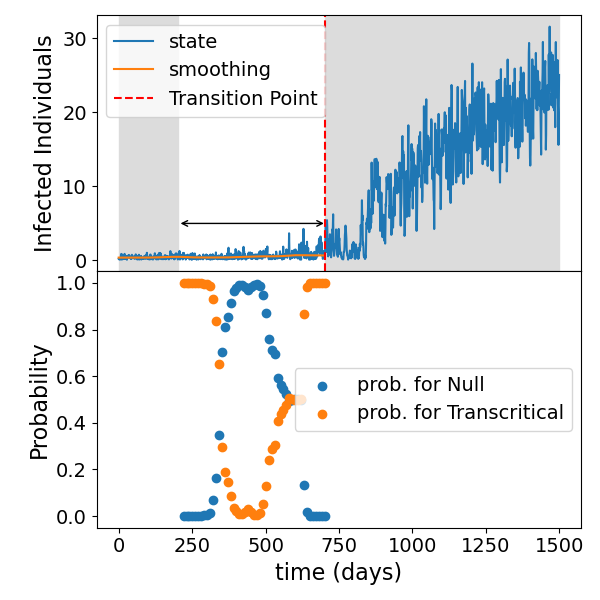}
 \end{subfigure}
  \hfill 
 \begin{subfigure}{0.19\textwidth}
     \includegraphics[width=\textwidth]{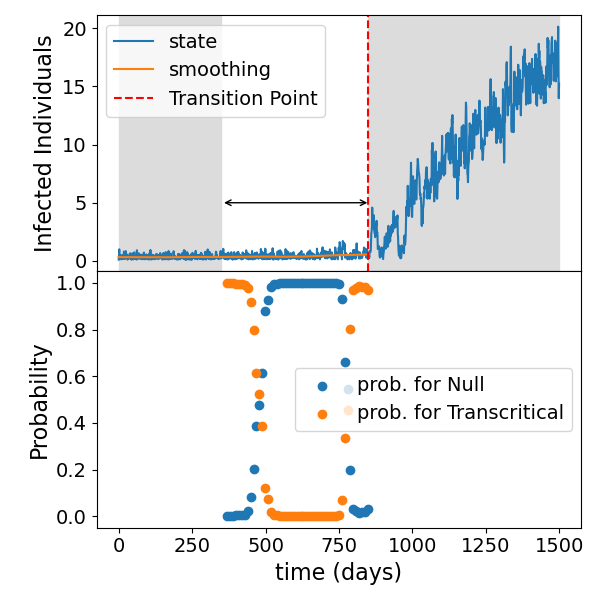}
 \end{subfigure} 
\medskip
  \begin{subfigure}[t]{0.19\textwidth}
    \includegraphics[width=\textwidth]{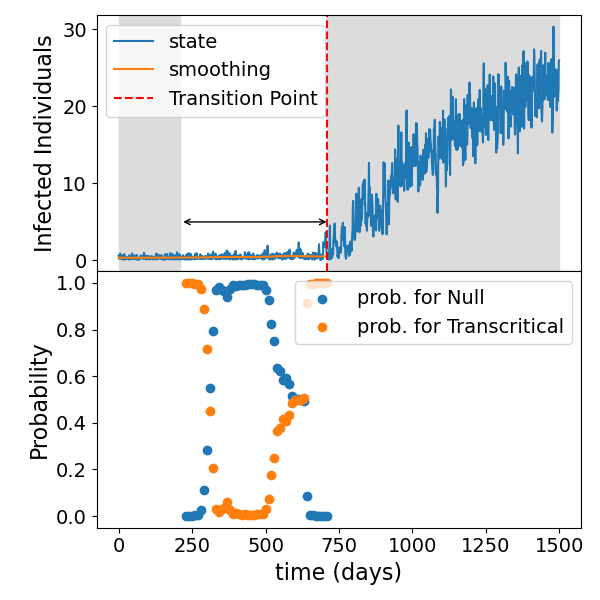}
 \end{subfigure}
  \hfill 
 \begin{subfigure}[t]{0.19\textwidth}
    \includegraphics[width=\textwidth]{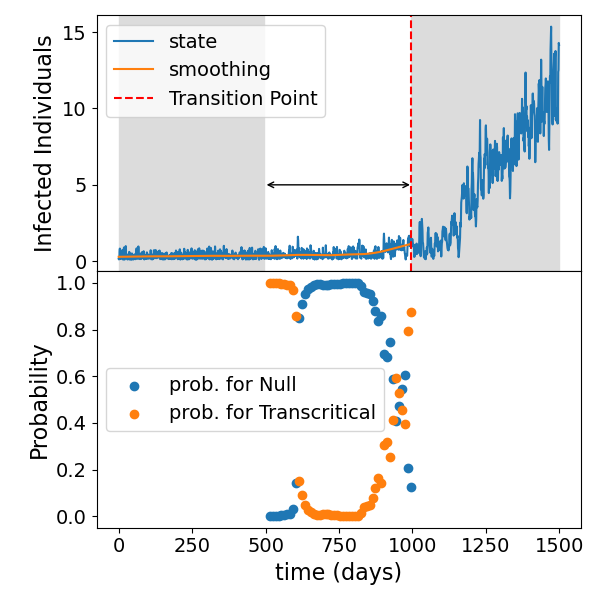}
 \end{subfigure}
  \hfill 
 \begin{subfigure}{0.19\textwidth}
     \includegraphics[width=\textwidth]{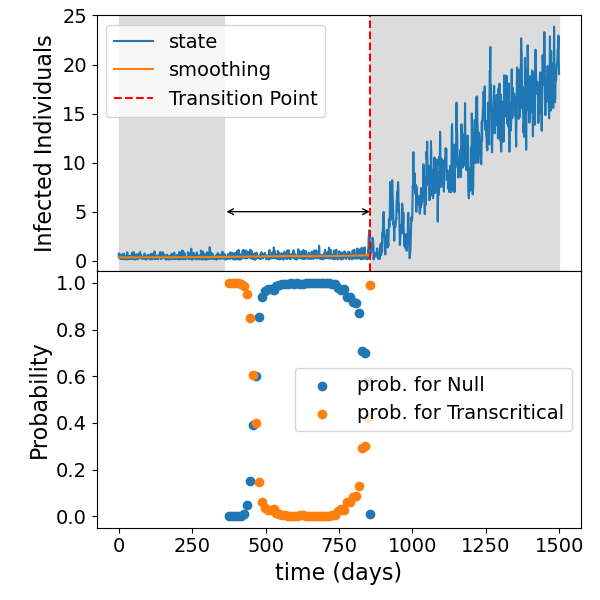}
 \end{subfigure}
  \hfill 
  \begin{subfigure}[t]{0.19\textwidth}
    \includegraphics[width=\textwidth]{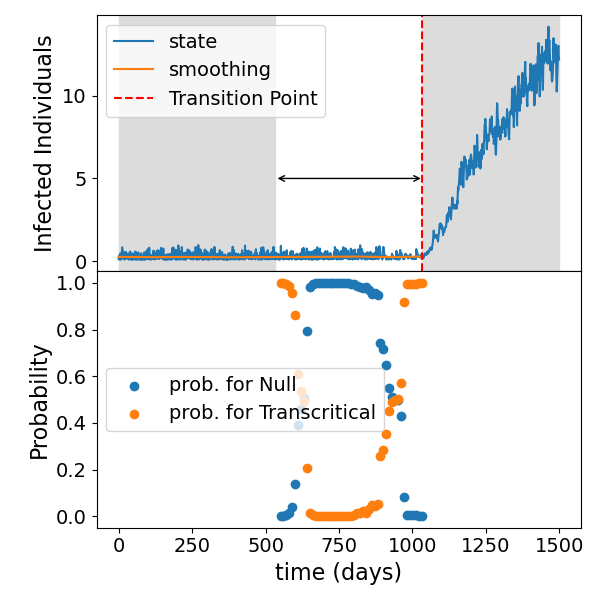}
 \end{subfigure}
  \hfill 
 \begin{subfigure}{0.19\textwidth}
    \includegraphics[width=\textwidth]{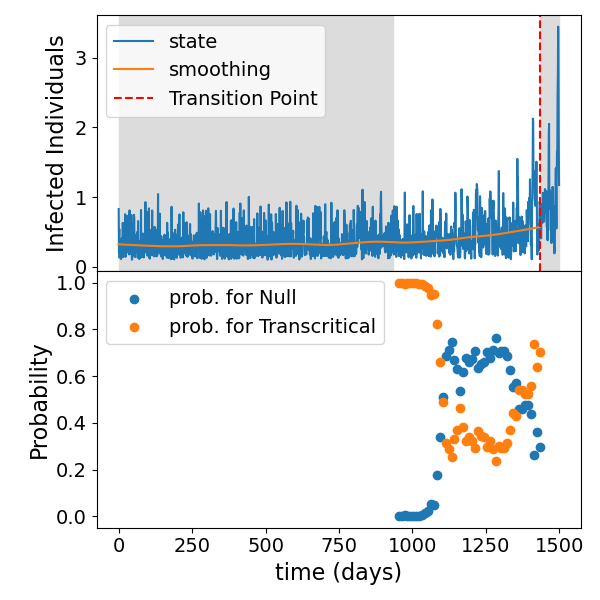}
 \end{subfigure}
\caption{Probabilities for a transition assigned by the SIDATR-500 DL model based on the subset of observations of transcritical simulations of the SIR model with demographic noise.}
\label{prob_dem_noise_trans}
\end{figure}

\begin{figure}
\centering
 \begin{subfigure}[t]{0.19\textwidth}
    \includegraphics[width=\textwidth]{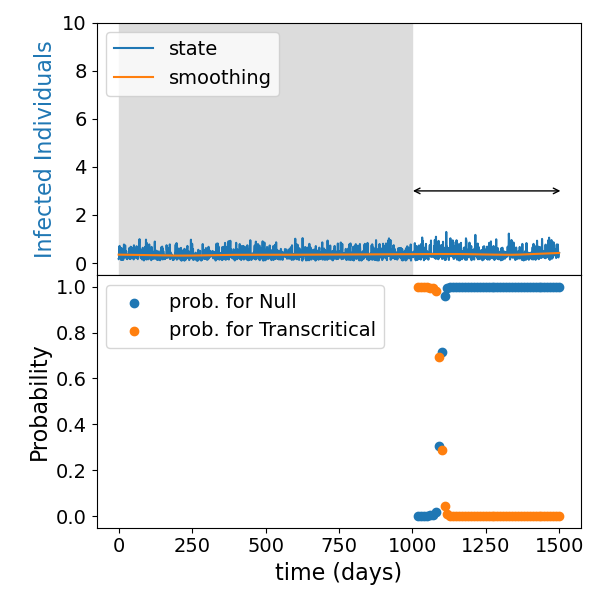}
 \end{subfigure}
 \hfill 
 \begin{subfigure}[t]{0.19\textwidth}
    \includegraphics[width=\textwidth]{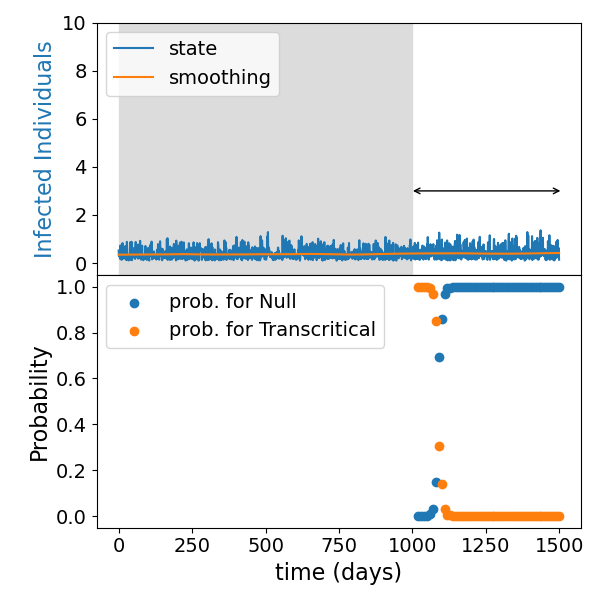}
 \end{subfigure}
  \hfill 
 \begin{subfigure}{0.19\textwidth}
     \includegraphics[width=\textwidth]{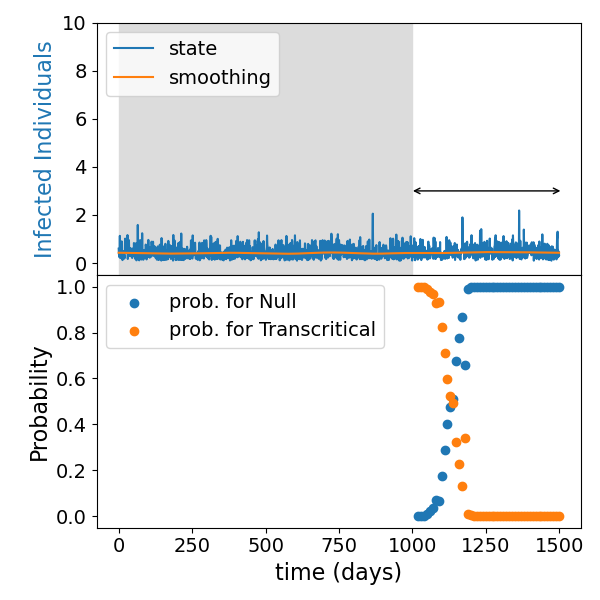}
 \end{subfigure}
  \hfill 
  \begin{subfigure}[t]{0.19\textwidth}
    \includegraphics[width=\textwidth]{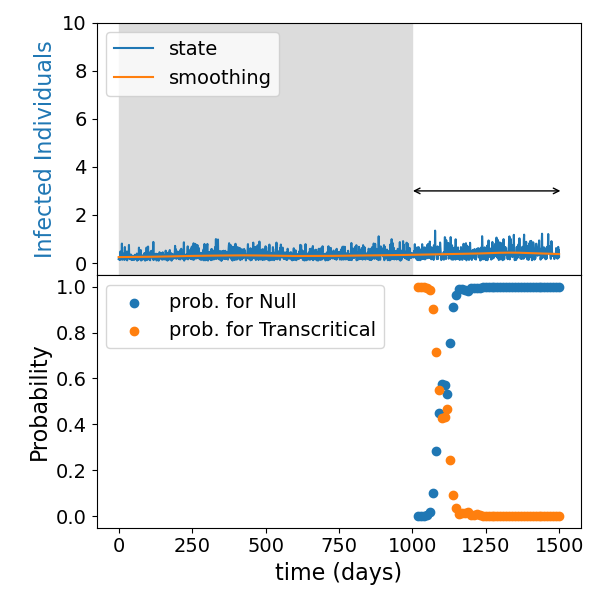}
 \end{subfigure}
  \hfill 
 \begin{subfigure}{0.19\textwidth}
     \includegraphics[width=\textwidth]{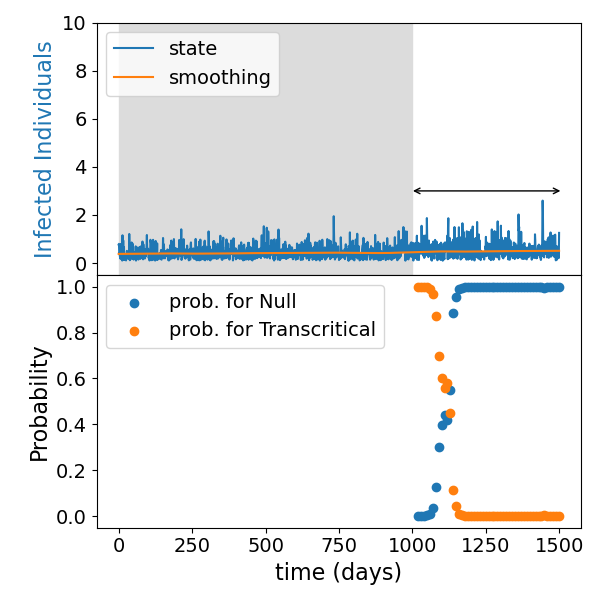}
 \end{subfigure} 
\medskip
  \begin{subfigure}[t]{0.19\textwidth}
    \includegraphics[width=\textwidth]{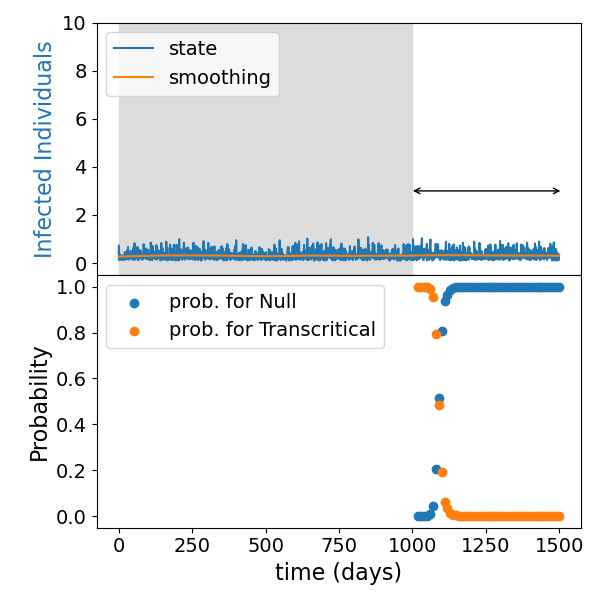}
 \end{subfigure}
  \hfill 
 \begin{subfigure}[t]{0.19\textwidth}
    \includegraphics[width=\textwidth]{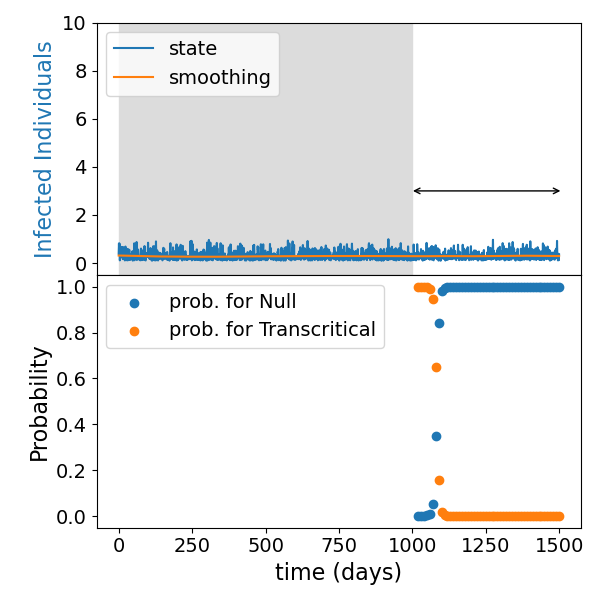}
 \end{subfigure}
  \hfill 
 \begin{subfigure}{0.19\textwidth}
     \includegraphics[width=\textwidth]{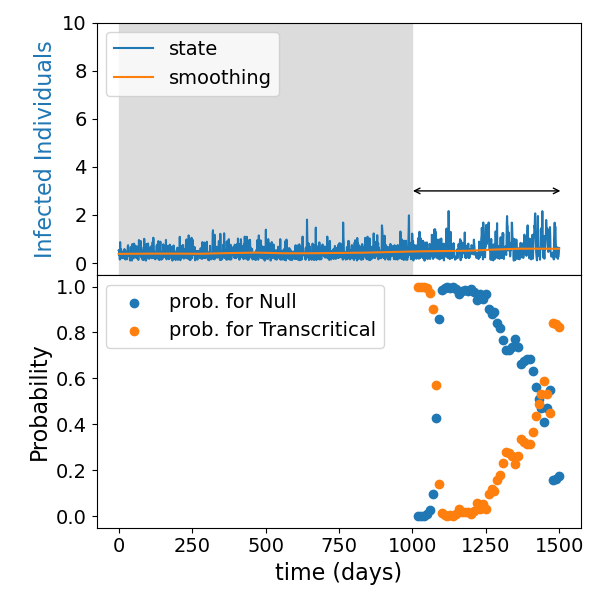}
 \end{subfigure}
  \hfill 
  \begin{subfigure}[t]{0.19\textwidth}
    \includegraphics[width=\textwidth]{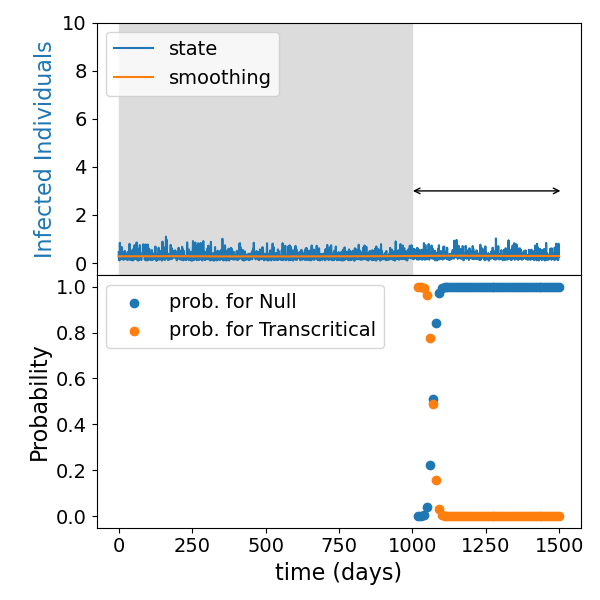}
 \end{subfigure}
  \hfill 
 \begin{subfigure}{0.19\textwidth}
    \includegraphics[width=\textwidth]{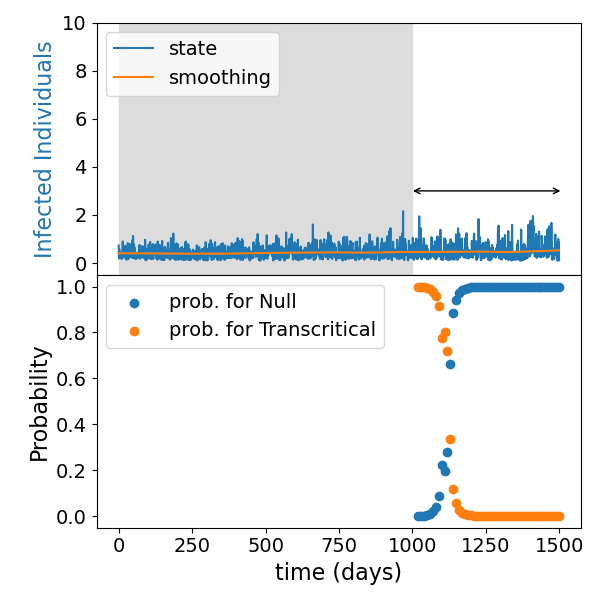}
 \end{subfigure}
\caption{Probabilities for a transition assigned by the SIDATR-500 DL model based on the subset of observations of null simulations of the SIR model with demographic noise.}
\label{prob_dem_noise_null}
\end{figure}

\begin{figure}
\centering
 \begin{subfigure}[t]{0.19\textwidth}
    \includegraphics[width=\textwidth]{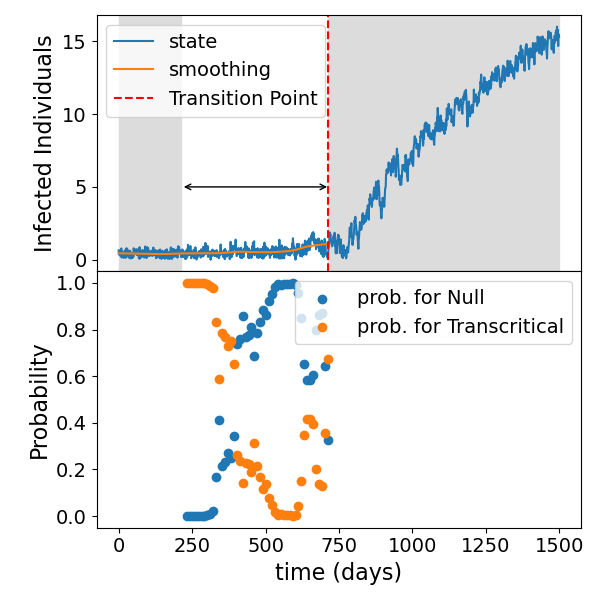}
 \end{subfigure}
 \hfill 
 \begin{subfigure}[t]{0.19\textwidth}
    \includegraphics[width=\textwidth]{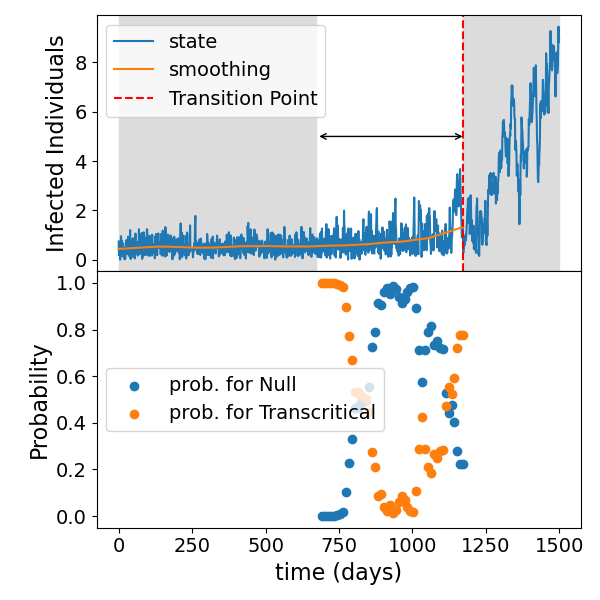}
 \end{subfigure}
  \hfill 
 \begin{subfigure}{0.19\textwidth}
     \includegraphics[width=\textwidth]{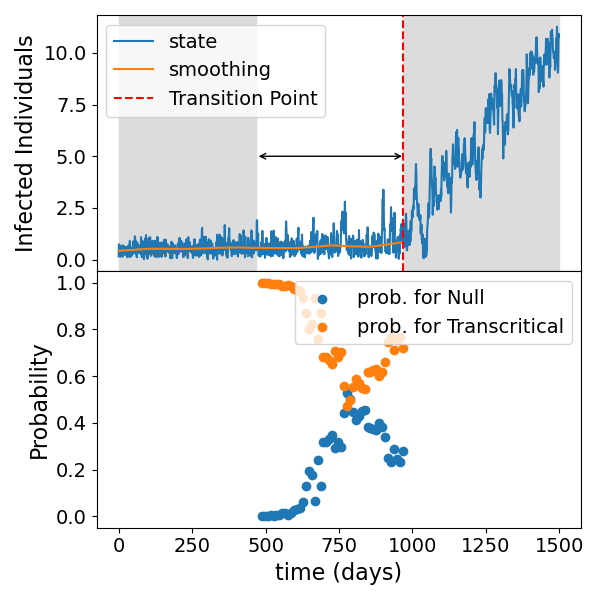}
 \end{subfigure}
  \hfill 
  \begin{subfigure}[t]{0.19\textwidth}
    \includegraphics[width=\textwidth]{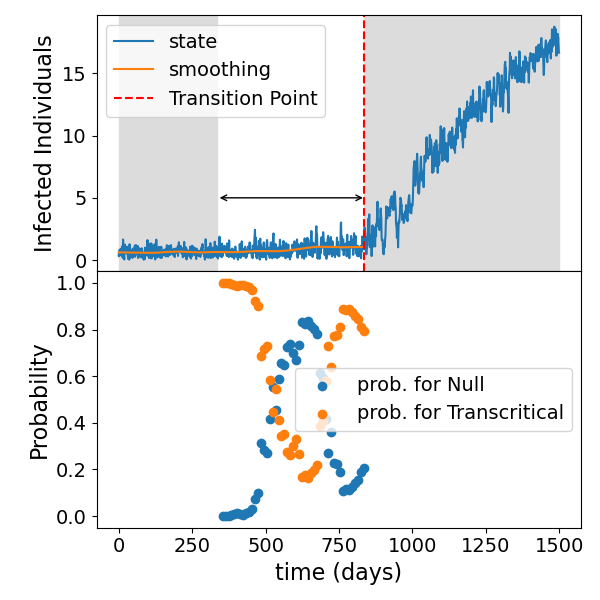}
 \end{subfigure}
  \hfill 
 \begin{subfigure}{0.19\textwidth}
     \includegraphics[width=\textwidth]{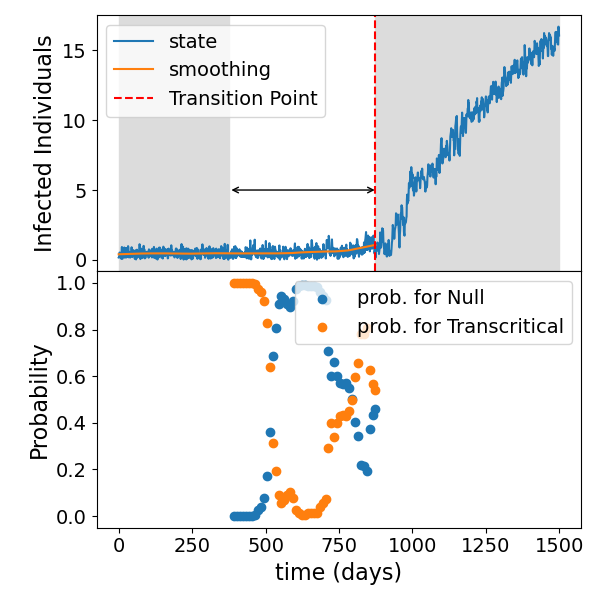}
 \end{subfigure} 
\medskip
  \begin{subfigure}[t]{0.19\textwidth}
    \includegraphics[width=\textwidth]{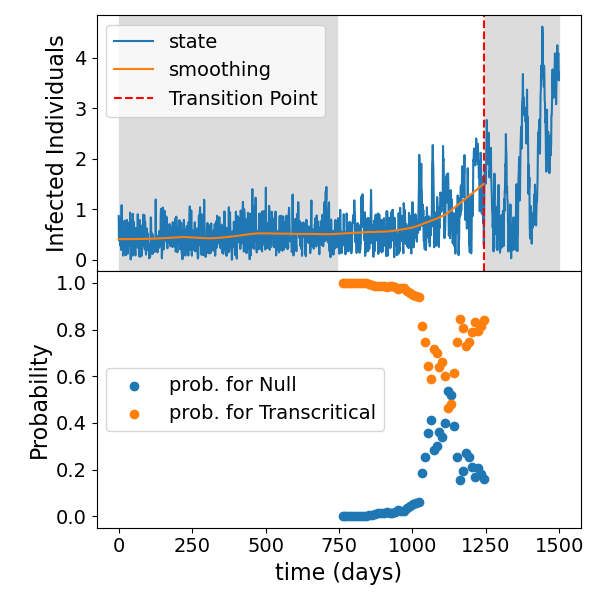}
 \end{subfigure}
  \hfill 
 \begin{subfigure}[t]{0.19\textwidth}
    \includegraphics[width=\textwidth]{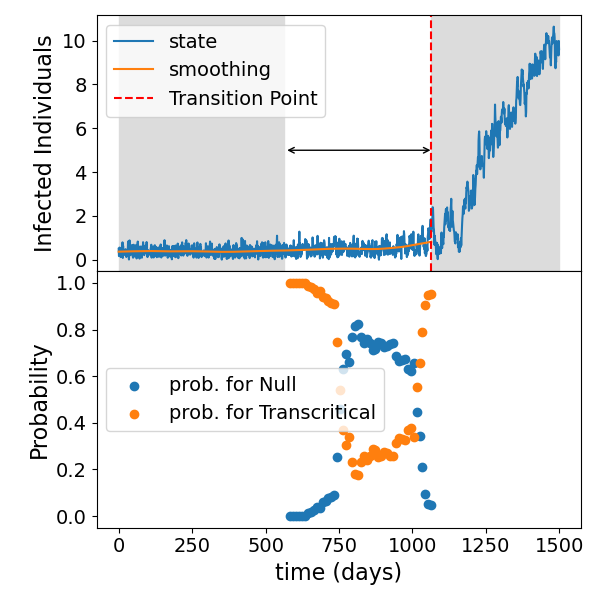}
 \end{subfigure}
  \hfill 
 \begin{subfigure}{0.19\textwidth}
     \includegraphics[width=\textwidth]{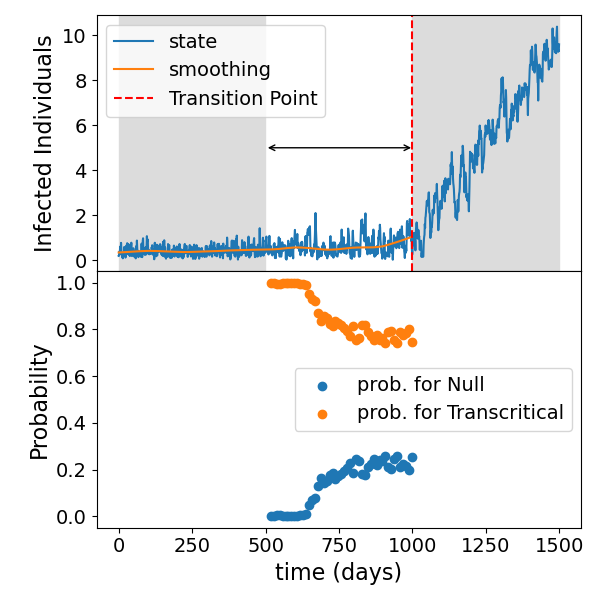}
 \end{subfigure}
  \hfill 
  \begin{subfigure}[t]{0.19\textwidth}
    \includegraphics[width=\textwidth]{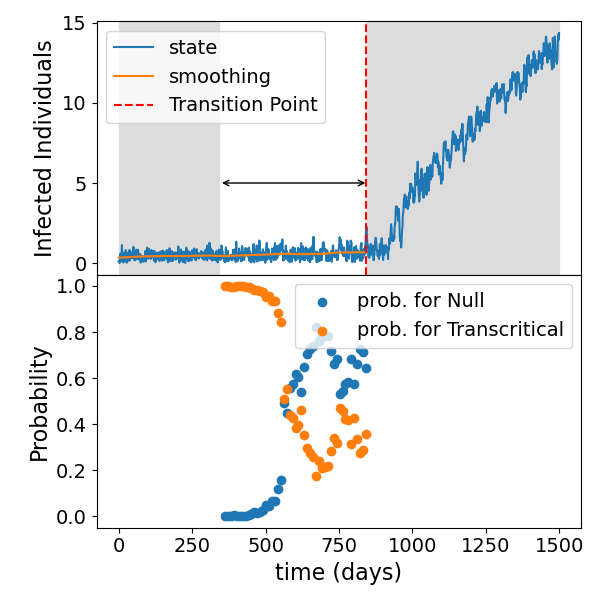}
 \end{subfigure}
  \hfill 
 \begin{subfigure}{0.19\textwidth}
    \includegraphics[width=\textwidth]{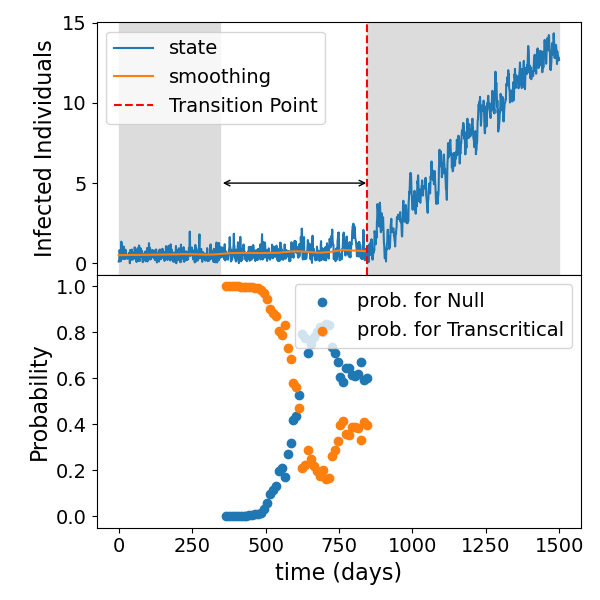}
 \end{subfigure}
\caption{Probabilities for a transition assigned by the SIDATR-500 DL model based on the subset of observations of transcritical simulations of the SEIR model with additive white noise.}
\label{prob_SEIR_trans}
\end{figure}

\begin{figure}
\centering
 \begin{subfigure}[t]{0.19\textwidth}
    \includegraphics[width=\textwidth]{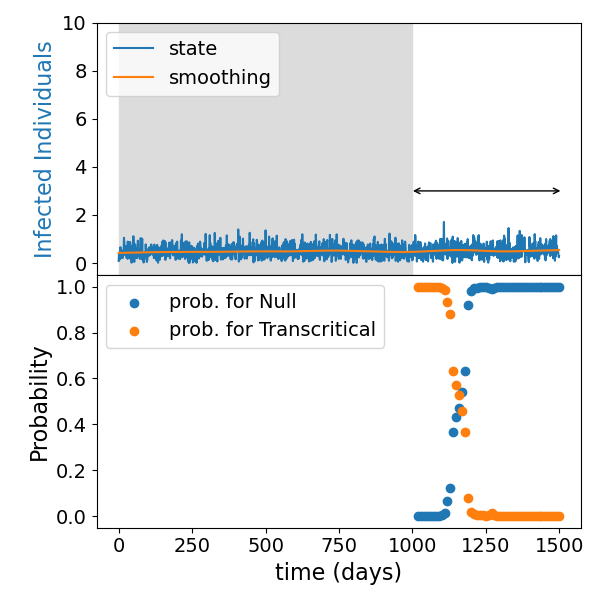}
 \end{subfigure}
 \hfill 
 \begin{subfigure}[t]{0.19\textwidth}
    \includegraphics[width=\textwidth]{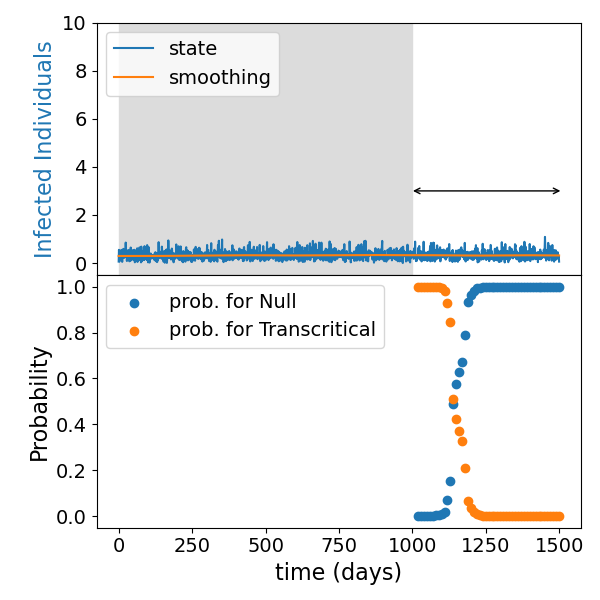}
 \end{subfigure}
  \hfill 
 \begin{subfigure}{0.19\textwidth}
     \includegraphics[width=\textwidth]{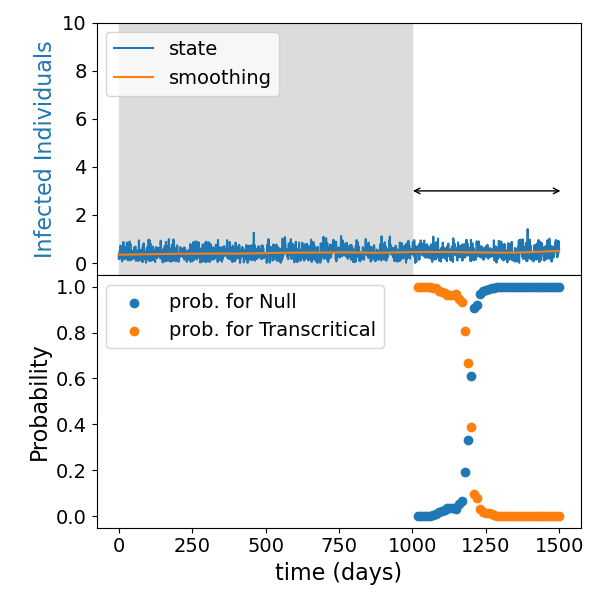}
 \end{subfigure}
  \hfill 
  \begin{subfigure}[t]{0.19\textwidth}
    \includegraphics[width=\textwidth]{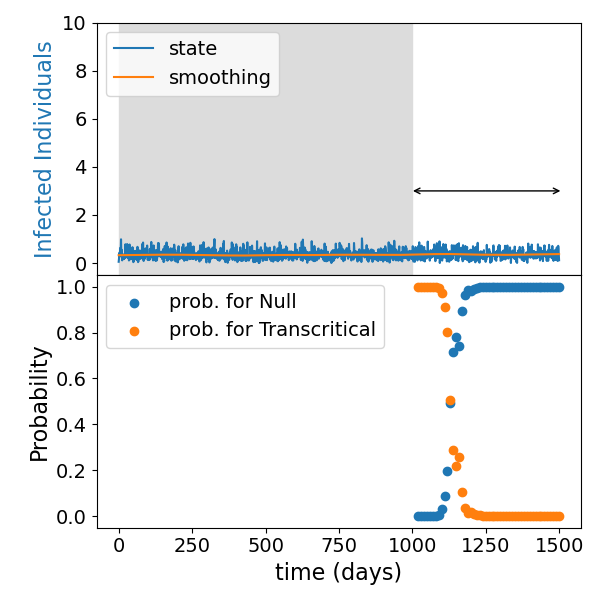}
 \end{subfigure}
  \hfill 
 \begin{subfigure}{0.19\textwidth}
     \includegraphics[width=\textwidth]{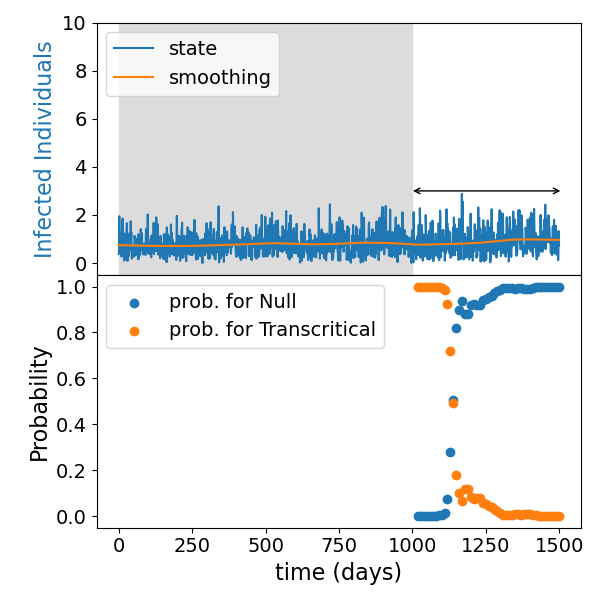}
 \end{subfigure} 
\medskip
  \begin{subfigure}[t]{0.19\textwidth}
    \includegraphics[width=\textwidth]{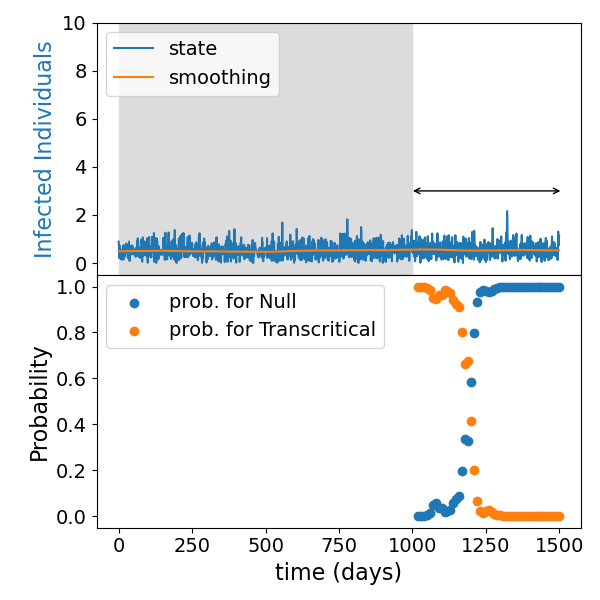}
 \end{subfigure}
  \hfill 
 \begin{subfigure}[t]{0.19\textwidth}
    \includegraphics[width=\textwidth]{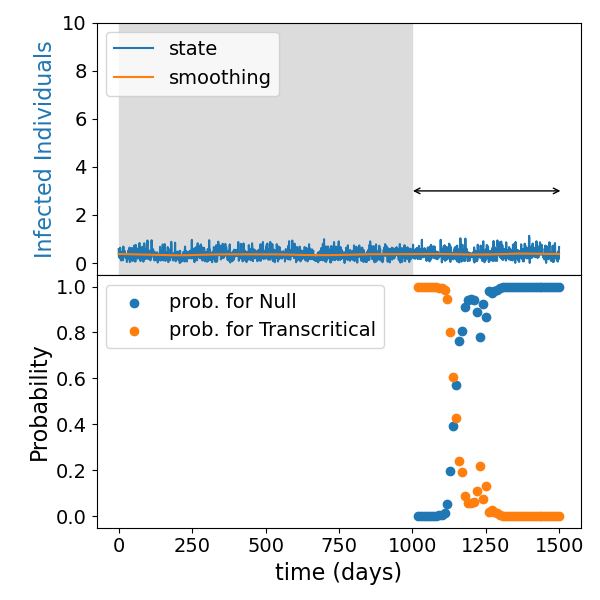}
 \end{subfigure}
  \hfill 
 \begin{subfigure}{0.19\textwidth}
     \includegraphics[width=\textwidth]{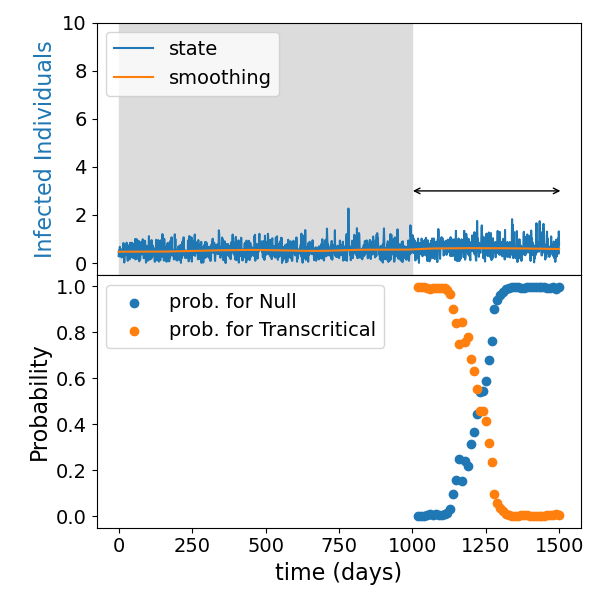}
 \end{subfigure}
  \hfill 
  \begin{subfigure}[t]{0.19\textwidth}
    \includegraphics[width=\textwidth]{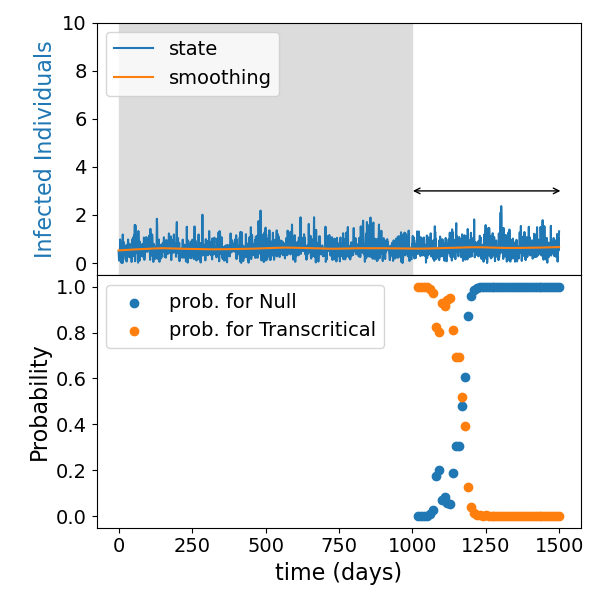}
 \end{subfigure}
  \hfill 
 \begin{subfigure}{0.19\textwidth}
    \includegraphics[width=\textwidth]{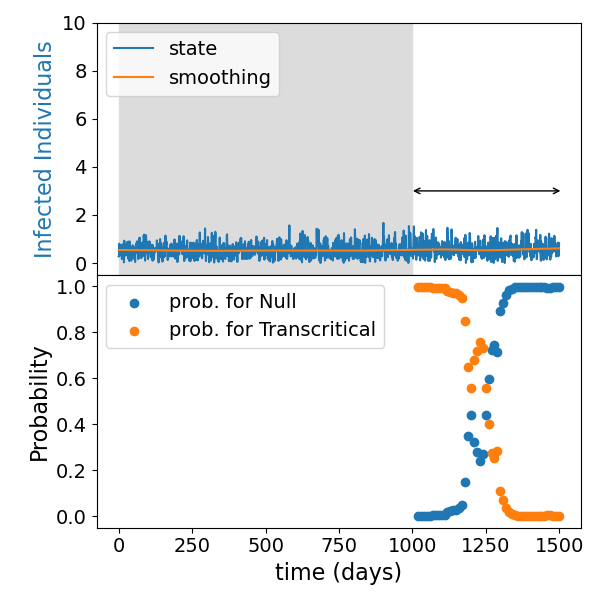}
 \end{subfigure}
\caption{Probabilities for a transition assigned by the SIDATR-500 DL model based on the subset of observations of null simulations of the SEIR model with additive white noise.}
\label{prob_SEIR_null}
\end{figure}

\begin{figure}
\centering
 \begin{subfigure}[t]{0.24\textwidth}
    \includegraphics[width=\textwidth]{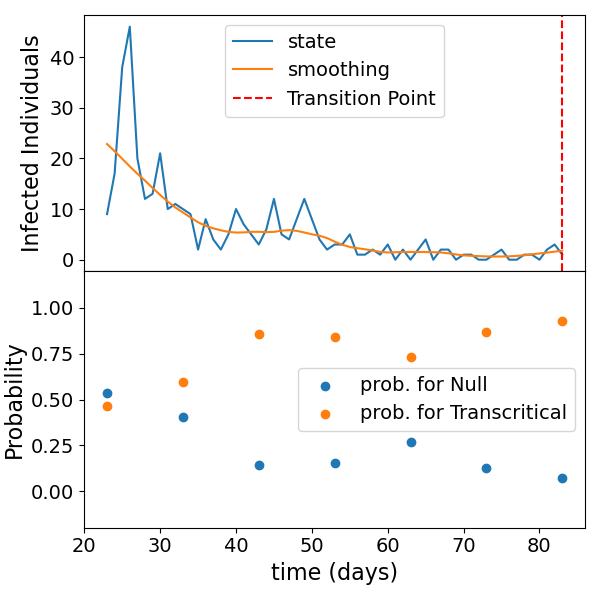}
 \end{subfigure}
 \hfill 
 \begin{subfigure}[t]{0.24\textwidth}
    \includegraphics[width=\textwidth]{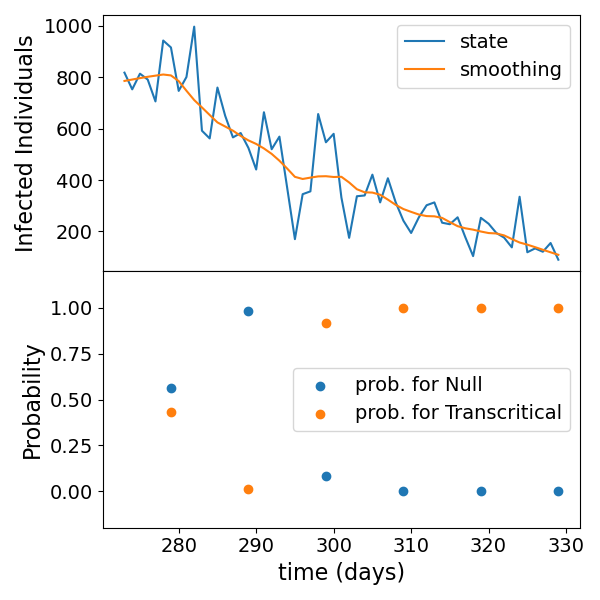}
 \end{subfigure}
  \hfill 
 \begin{subfigure}{0.24\textwidth}
     \includegraphics[width=\textwidth]{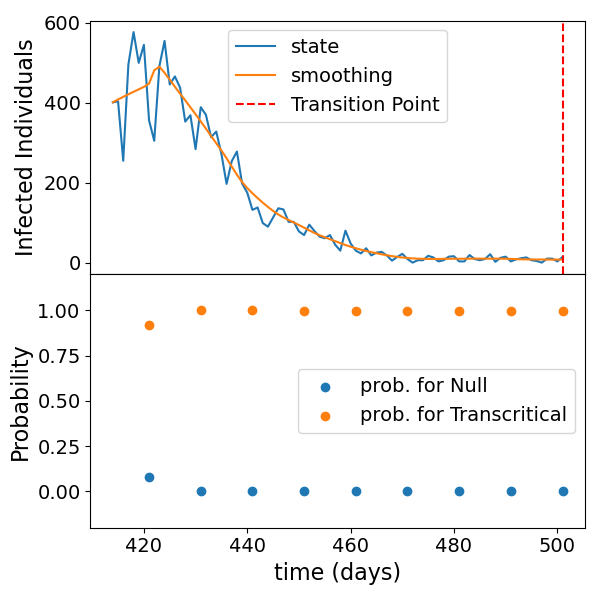}
 \end{subfigure}
  \hfill 
  \begin{subfigure}[t]{0.24\textwidth}
    \includegraphics[width=\textwidth]{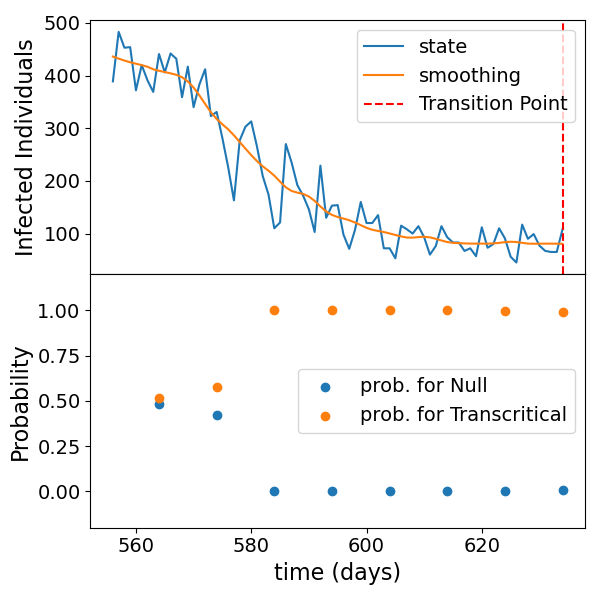}
 \end{subfigure}
  \hfill 
\medskip
 \begin{subfigure}{0.24\textwidth}
     \includegraphics[width=\textwidth]{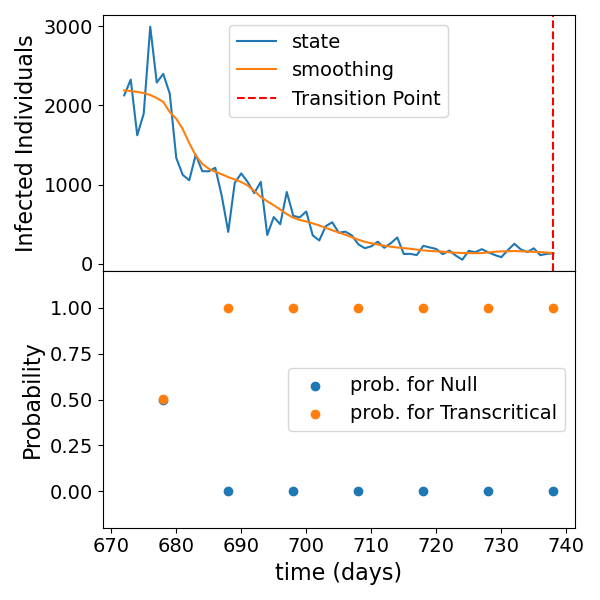}
 \end{subfigure} 
  \begin{subfigure}[t]{0.24\textwidth}
    \includegraphics[width=\textwidth]{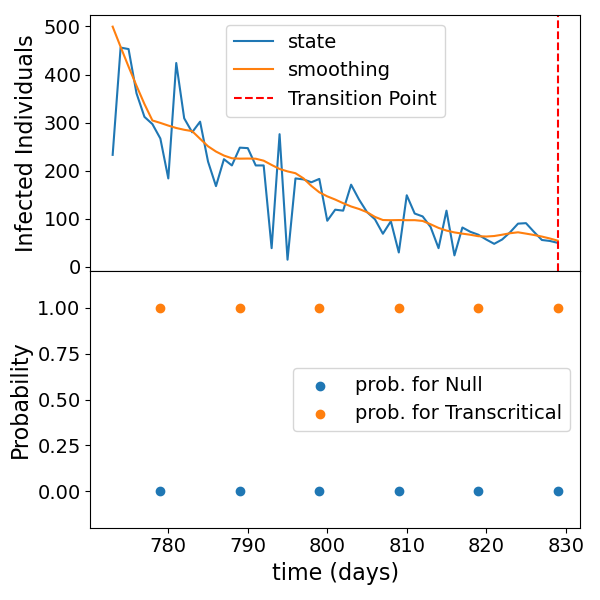}
 \end{subfigure}
  \hfill 
 \begin{subfigure}[t]{0.24\textwidth}
    \includegraphics[width=\textwidth]{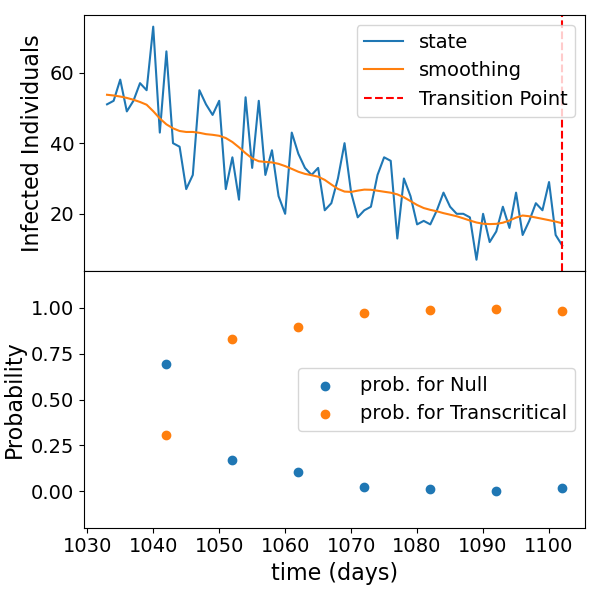}
 \end{subfigure}
\caption{Probabilities for a transition assigned by the SIDATR-100 DL model based on the transcritical time series of the COVID-19 dataset of Edmonton.}
\label{prob_COVID_ED_for}
\end{figure}

\begin{figure}
\centering
 \begin{subfigure}[t]{0.24\textwidth}
    \includegraphics[width=\textwidth]{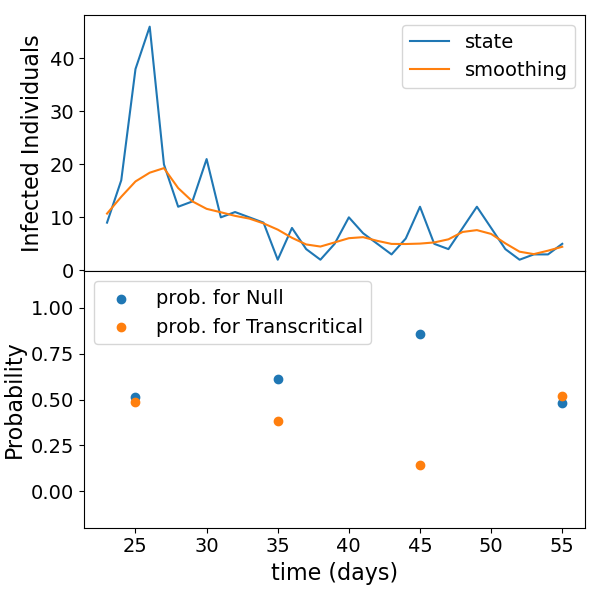}
 \end{subfigure}
 \hfill 
 \begin{subfigure}[t]{0.24\textwidth}
    \includegraphics[width=\textwidth]{figures/COVID_null_plot2.png}
 \end{subfigure}
  \hfill 
 \begin{subfigure}{0.24\textwidth}
     \includegraphics[width=\textwidth]{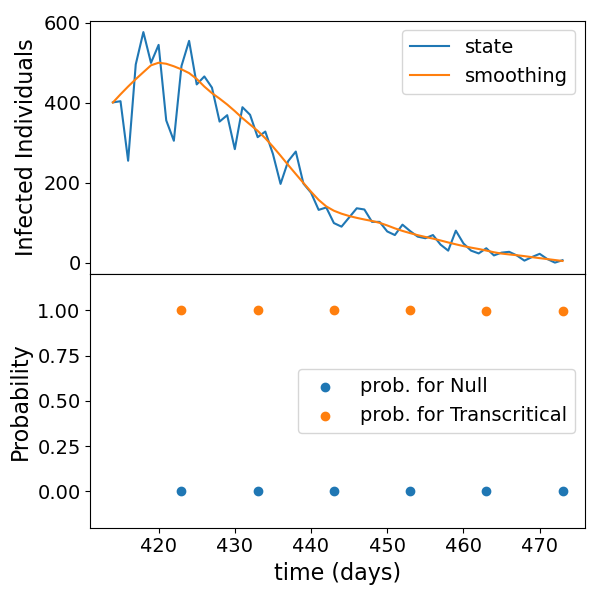}
 \end{subfigure}
  \hfill 
  \begin{subfigure}[t]{0.24\textwidth}
    \includegraphics[width=\textwidth]{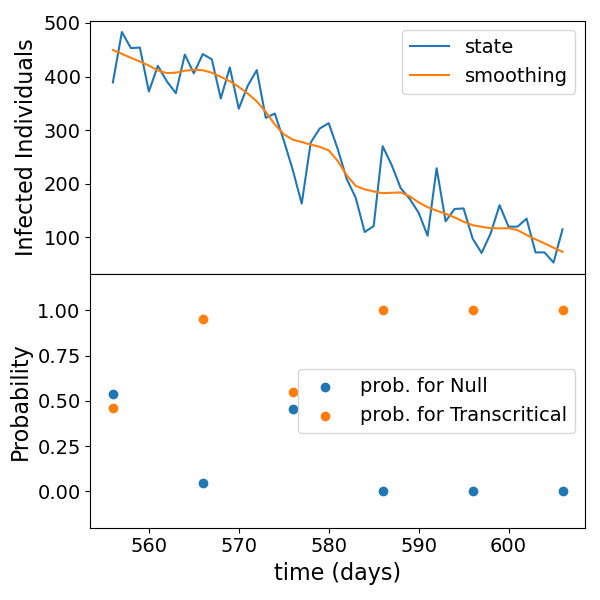}
 \end{subfigure}
  \hfill  
\medskip
 \begin{subfigure}{0.24\textwidth}
     \includegraphics[width=\textwidth]{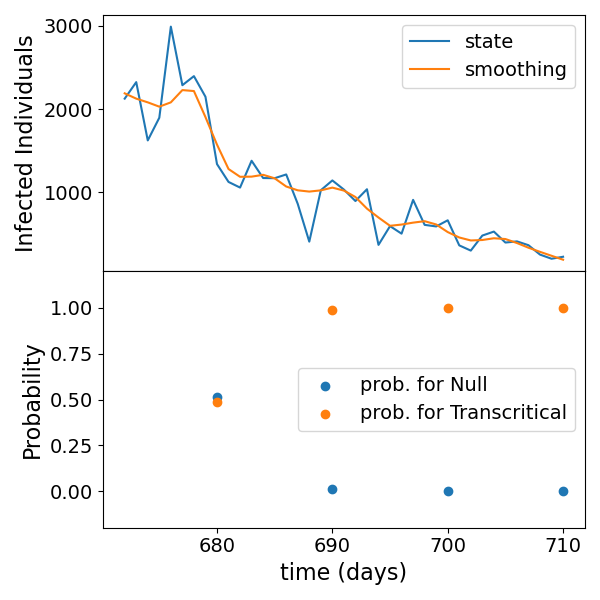}
 \end{subfigure}
  \begin{subfigure}[t]{0.24\textwidth}
    \includegraphics[width=\textwidth]{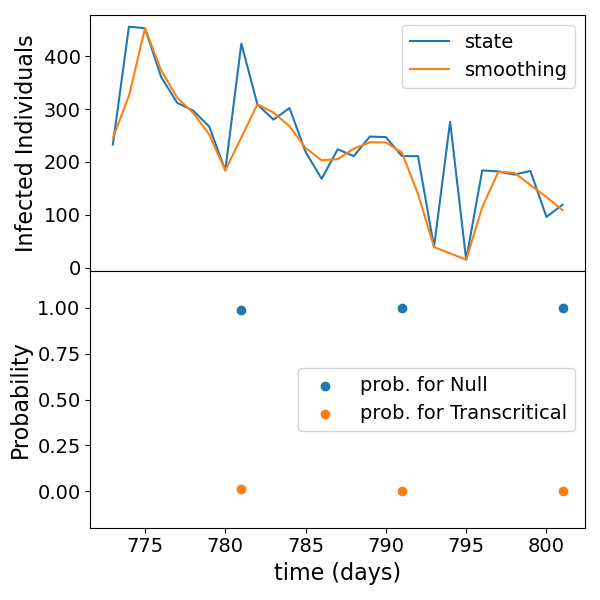}
 \end{subfigure}
  \hfill 
 \begin{subfigure}[t]{0.24\textwidth}
    \includegraphics[width=\textwidth]{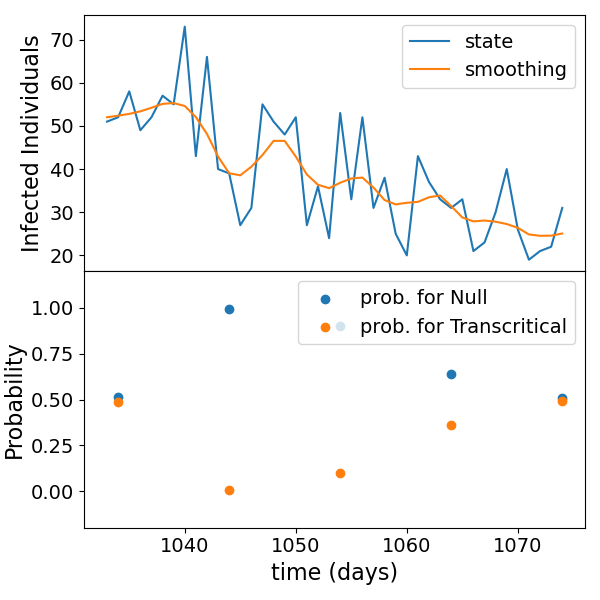}
 \end{subfigure}
\caption{Probabilities for a transition assigned by the SIDATR-100 DL model based on the null time series of the COVID-19 dataset of Edmonton.}
\label{prob_COVID_ED_null}
\end{figure}